\documentclass[final,1p,times]{elsarticle}

\usepackage{amssymb}
\usepackage{amsmath}
\usepackage{subfigure}
\usepackage{multirow}
\usepackage{url}
\usepackage{tabularx}
\usepackage{bbding}
\usepackage{graphicx}
\usepackage{makecell}
\usepackage[figuresright]{rotating}
\usepackage{tablefootnote}
\usepackage{threeparttable}
\usepackage[normalem]{ulem}

\usepackage{appendix}
\usepackage[colorlinks,bookmarksopen,bookmarksnumbered,citecolor=blue, linkcolor=blue, urlcolor=blue]{hyperref}
\usepackage{color}

\usepackage{rotating}  
\usepackage{hyperref} 
\usepackage{booktabs}

\begin{document}

\begin{frontmatter}



\title{Reviewing Clinical Knowledge in Medical Large Language Models: Training and Beyond}


\author[label1,label2]{Qiyuan Li}
\ead{vickyuan@wust.edu.cn}
\author[label1,label2]{Haijiang Liu}
\ead{alecliu@ontoweb.wust.edu.cn}
\author[label1,label2]{Caicai Guo}
\ead{guocaicai@ontoweb.wust.edu.cn}
\author[label1]{Chao Gao}
\ead{gchao@ieee.org}
\author[label4]{Deyu Chen}
\ead{deyuchen@hust.edu.cn}
\author[label5]{Meng Wang}
\ead{wang_meng@whu.edu.cn}
\author[label1,label2]{Feng Gao}
\ead{feng.gao86@wust.edu.cn}
\author[label6]{Frank van Harmelen}
\ead{frank.van.harmelen@vu.nl}

\author[label1,label2]{Jinguang Gu\corref{authorinfo}}
\ead{simon@wust.edu.cn}
\cortext[authorinfo]{Corresponding author at: School of Computer Science and Technology, Wuhan University of Science and Technology.}


\affiliation[label1]{organization={School of Computer Science and Technology, Wuhan University of Science and Technology},
            city={Wuhan},
            postcode={430065}, 
            state={Hubei},
            country={China}}
\affiliation[label2]{organization={Hubei Province Key Laboratory of Intelligent Information Processing and Real-time Industrial System},
            city={Wuhan},
            postcode={430065}, 
            state={Hubei},
            country={China}}

\affiliation[label4]{organization={School of Computer Science and Technology, Huazhong University of Science and Technology},
            city={Wuhan},
            postcode={430074}, 
            state={Hubei},
            country={China}}

\affiliation[label5]{organization={School of Cyber Science and Engineering, Wuhan University},
            city={Wuhan},
            postcode={430072}, 
            state={Hubei},
            country={China}}

\affiliation[label6]{organization={Department of Computer Science, Vrije Universiteit Amsterdam},
            city={Amsterdam},
            country={The Netherlands}}

\begin{abstract}
The large-scale development of large language models (LLMs) in medical contexts, such as diagnostic assistance and treatment recommendations, necessitates that these models possess accurate medical knowledge and deliver traceable decision-making processes. Clinical knowledge, encompassing the insights gained from research on the causes, prognosis, diagnosis, and treatment of diseases, has been extensively examined within real-world medical practices. Recently, there has been a notable increase in research efforts aimed at integrating this type of knowledge into LLMs, encompassing not only traditional text and multimodal data integration but also technologies such as knowledge graphs (KGs) and retrieval-augmented generation (RAG). In this paper, we review the various initiatives to embed clinical knowledge into training-based, KG-supported, and RAG-assisted LLMs. We begin by gathering reliable knowledge sources from the medical domain, including databases and datasets. Next, we evaluate implementations for integrating clinical knowledge through specialized datasets and collaborations with external knowledge sources such as KGs and relevant documentation. Furthermore, we discuss the applications of the developed medical LLMs in the industrial sector to assess the disparity between models developed in academic settings and those in industry. We conclude the survey by presenting evaluation systems applicable to relevant tasks and identifying potential challenges facing this field. In this review, we do not aim for completeness, since any ostensibly "complete" review would soon be outdated. Our goal is to illustrate diversity by selecting representative and accessible items from current research and industry practices, reflecting real-world situations rather than claiming completeness. Thus, we emphasize showcasing diverse approaches.
\end{abstract}




\end{frontmatter}


\section{Introduction}
\label{}
Clinical knowledge, derived from a large amount of medical literature, expert experience, and cases, is the principles doctors follow in diagnosis and treatment. Various studies have integrated clinical knowledge in accurate medical artificial intelligence (AI) systems for real-life assistance. Traditional medical AI research \cite{Jan22Medical, Andrés21an} tends to discover patterns from large amounts of medical data to learn these principles. With the development of deep machine learning, two lines of studies have formed to build a more accurate model: direct learning \cite{Farzan23Deep} and knowledge extraction \cite{Enayat22Knowledge, Linfeng20Real}. 

(1) Researchers on direct learning \cite{Farzan23Deep} use deep neural networks to learn knowledge directly from massive medical data and build medical models based on clinical knowledge. Integrating and modeling this knowledge can provide a deeper understanding of the disease mechanism and support clinical decision-making. 
(2) Researchers on knowledge extractions \cite{Linfeng20Real} have developed natural language processing (NLP) frameworks to extract knowledge from vast medical corpora and construct clinical medical knowledge graphs (KGs), which are used as external knowledge to assist prediction models. Moreover, these clinical knowledge bases (KBs) are continuously improved and updated to help medical models better adapt to clinical needs and improve diagnosis and treatment. 

Large language models (LLMs) and their powerful learning capabilities have recently attracted attention in medical AI \cite{Thirunavukarasu23Large, Wornow23TheShaky, Safranek23TheRole, Zhou23ASurvey, He23ASurvey, Ali23Deep, Yanjun23Leveraging}. By mining massive medical data, such as electronic medical records (EMRs), medical images, and genetic information, these models aim to learn more accurate and fine-grained knowledge in medical data and provide more efficient diagnostics for doctors in real-life medical scenarios. Incorporating clinical knowledge into medical AI can play crucial roles in real-world medical scenarios:
\begin{itemize}
    \item Improve models' development and capabilities in the medical field. By analyzing a large amount of clinical data, medical LLMs can continuously adjust and optimize model parameters to improve the accuracy of prediction and diagnosis. At the same time, clinical knowledge can also guide researchers to pay attention to the differences and specificities of the patient population during model training and improve the model's generalization ability. This helps medical AI systems play a greater role in clinical applications and improve the quality of healthcare services.
    \item Improve the diagnostic efficiency of real-life clinical practice. Medical LLMs can use clinical knowledge to quickly and accurately diagnose the patient's condition and provide doctors with treatment plans. In clinical practice, medical AI systems can recommend personalized treatment plans based on the patient's condition, present history, family history, and other information to improve the treatment. In addition, medical LLMs can also discover potential disease risks and complications by mining and analyzing clinical data, providing the basis for clinical prevention and management.
\end{itemize}

So far, there are numerous comprehensive reviews of medical AI. For instance, Thirunavukarasu et al. \cite{Thirunavukarasu23Large} focuses on the efficiency and effectiveness of LLM applications, such as the power of ChatGPT \cite{ChatGPT} in clinical, educational, and research. Wornow et al. \cite{Wornow23TheShaky} investigates 84 base models in non-imaging EMR data and proposes an enhanced evaluation framework for these models to align with essential metrics in healthcare closely. Safranek et al. \cite{Safranek23TheRole} highlights the critical limitations of LLM in medical education. Zhou et al. \cite{Zhou23ASurvey} summarises the approaches to constructing medical LLMs, compares the performance of existing models, and provides relevant insights into the development opportunities. He et al. \cite{He23ASurvey} outline the development process of the LLM created by healthcare and provide an overview of the development roadmap from the traditional pre-trained language model (PLM) to the LLM. Ali et al. \cite{Ali23Deep} reported recent advances in deep learning-based drug recommendation methods.
Based on the above-detailed research, we present a comprehensive review of data, methodology, evaluation, and application in the clinical literature, as well as explore the differences between academic research and industry practices of building medical AI systems.
 
Therefore, this paper begins by discussing existing clinical databases and datasets. Based on this, we elaborate on the construction mechanisms of medical academic LLMs and analyze the challenges of current medical model research. In addition to analyzing medical LLMs trained on clinical databases and datasets, we have also presented the integration of medical LLMs and KGs, which contain a rich and strong semantic connectivity. 
Then, we present an overview of practical applications in real-world medical circumstances and identify the gap between academic model research and industry practice. Additionally, we summarize the principles and methods for evaluating current medical LLMs and perform a systematic analysis. Finally, we propose potential future development directions for medical AI research, aiming to promote the deep integration of academic research and clinical practice and to provide scientific basis and practical guidance for the further development of medical AI.

\begin{figure}[htp]
    \centering
    \includegraphics[width=1\textwidth]{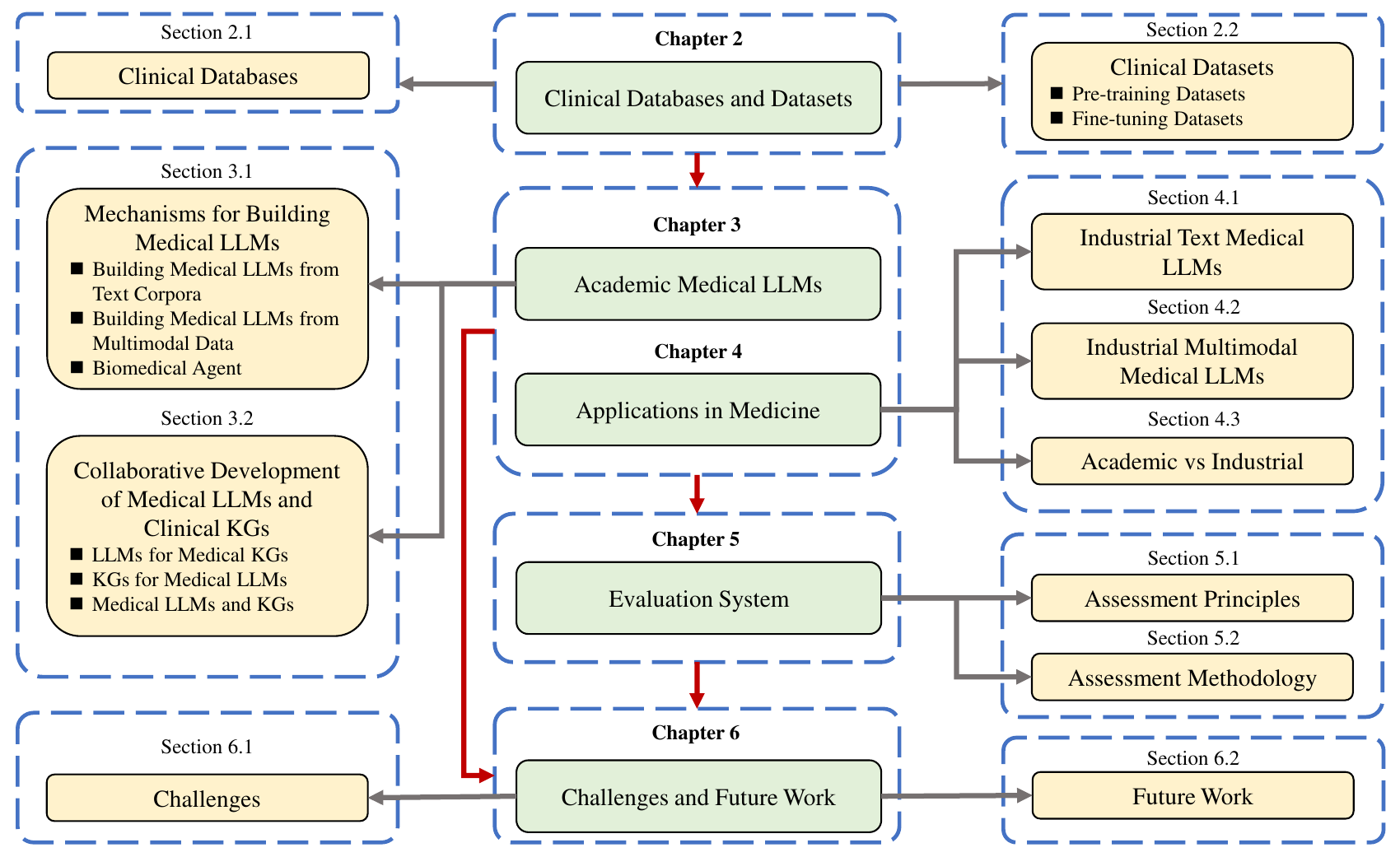}
    \caption{The overall structure of the paper. Chapter \ref{chp: 2}: clinical databases and datasets. Chapter \ref{chp: 3}: construction mechanisms of academic medical LLMs. Chapter \ref{chp: 4}: applications of medical LLMs. Chapter \ref{chp: 5}: evaluation of medical LLMs. Chapter \ref{chp: 6}: challenges and future works.}
    \label{fig:structure}
\end{figure}

The overall structure of this survey is shown in Figure \ref{fig:structure}. In the following chapters, we summarise existing clinical databases and datasets in Chapter \ref{chp: 2}. Subsequently, we analyze the construction mechanisms and performance of numerous academic medical LLMs in Chapter \ref{chp: 3}. Then, we list some of the medical LLMs that have attracted wide attention in practice and analyze the possible limitations of the current academic medical LLMs in Chapter \ref{chp: 4}. Furthermore, we systematically review the principles and methods for evaluating current medical LLMs and perform a systematic analysis in Chapter \ref{chp: 5}. Finally, we recommend how future scientific research on medical AI can better play a role in healthcare scenarios in Chapter \ref{chp: 6}.

\section{Clinical Databases and Datasets}
\label{chp: 2}
In this chapter, we will first introduce open-source clinical databases. Then, we will introduce the clinical datasets by categorizing them into pre-training and fine-tuning stages from the perspective of constructing medical LLMs. This will establish a solid data foundation for the next chapter, which focuses on building academic medical LLMs.

\subsection{Clinical Databases}
Clinical databases are comprehensive and systematic summaries of clinical knowledge, which contain a quantity of medical expertise and real-life clinical experience. Obtaining this knowledge can help LLMs identify deceases based on symptoms and assist in decision-making of complex conditions. LLMs utilizing clinical databases provide personalized treatment plans based on patient circumstances and make accurate treatment recommendations. More importantly, up-to-date clinical databases allow LLMs to update medical results and improve prediction qualities. The clinical databases serve as a bridge integrating clinical domain knowledge and lay a solid foundation for LLMs to learn domain knowledge. Table \ref{tbl: ckb} lists some commonly used clinical databases.

\begin{table*}[!htb]
    \footnotesize
	\caption{The collection of common clinical databases. "Text Corpora" indicates that the knowledge is stored as text, and "Multimodal" includes not only text but also other modalities such as images and videos.}
        \label{tbl: ckb}
	\centering
	\begin{tabularx}{\textwidth}{c X c}
        \hline
		\textbf{Name} & \centering \textbf{Content} & \textbf{Modality}  \\
        \hline
		Drugs & Drugs.com provides accurate and independent information on over 24,000 prescription drugs, over-the-counter medicines, and products. & Text Corpora 
		\\
        \hline
		DrugBank & The latest release of DrugBank Online contains 16,568 drug entries, including 2,761 approved small molecule drugs, 1,611 approved biologics, 135 nutraceuticals, and over 6,723 experimental drugs. Additionally, 5,303 non-redundant protein sequences are linked to these drug entries. & Text Corpora\\
        \hline
        NHS Medicine & NHS Medicine provides detailed information on 288 prescription drugs, over-the-counter medicines, drug-drug interaction, and side effects.& Text Corpora\\
        \hline
        Medline & The flagship bibliographic resource of the National Library of Medicine cites over 31 million journal articles across the life sciences, particularly on biomedical topics. & Text Corpora\\
        \hline
        Embase & Embase has over 32 million entries and over 2,900 proprietary journal indexes. Its unique Emtree thesaurus includes a MeSH thesaurus, 56,000 specialized search terms, and 230,000 synonyms. & Text Corpora\\
        \hline
        UMLS & UMLS includes about 2 million medical concepts and a medical vocabulary of more than 5 million words.& Text Corpora\\
        \hline
        HPO & HPO currently contains over 13,000 terms and over 156,000 annotations to hereditary diseases. & Text Corpora\\
        \hline
        UpToDate Clinical Advisor & UpToDate Clinical Advisor includes 12,400+ clinical topics covering 25 specialties, 9,800+ graded recommendations, 37,000+ image profiles, 220+ medical calculators, 7,600+ English-language drug monographs, and 544,000+ Medline references.& Multimodal\\
        \hline
        ClinicalKey & Clinicalkey comprises over 676 literature resources, 1,005 classic books, 63,699 medical videos, 4.64 million images, 5,000 clinical guidelines, 210,000 diagnostic trials, 2,562 drug monographs, 99.01 million patient educations, 50 North American clinics, 339 operating videos, 555 clinical spotlights, and 22 million Medline abstracts. & Multimodal\\
        \hline
        NHS Health & NHS Health provides an overview of 1,216 diseases, treatments, and information on healthy lifestyles.& Text Corpora\\
        \hline
	\end{tabularx}%
\end{table*}%

\paragraph{Structured Drug Information} Among them, drug information is one of the subjects that researchers focus on building KGs, as it is usually represented by structured descriptions, which make it easier to construct ontology and extract knowledge. This type of knowledge is often significant for clinical practice because of its informativeness. Common KBs for drug information are:
\begin{enumerate}
    \item \textbf{Drugs}\footnote{https://www.drugs.com/} provides a global online database of drug information, including descriptions, uses, dosages, side effects, and more. Users can access drug information for free, including details on drug names, specifications, manufacturers, and more. Additionally, it offers expert advice on drug interactions, side effects, and the use of drugs during pregnancy and breastfeeding.
    \item \textbf{DrugBank}\footnote{https://go.drugbank.com/} is an online database containing a vast amount of drug-related information. It is unique in that it provides information on the transformation process of drugs in the body, the mechanism of drug-drug interaction, and more. This makes it an invaluable resource for researchers and pharmaceutical professionals.
    \item \textbf{National Health Service (NHS) Medicine}\footnote{https://www.nhs.uk/medicines/} provides an online drug information platform that offers detailed information on prescription and over-the-counter drugs, drug-drug interaction, side effects, and more. The platform aims to help doctors, pharmacists, and patients understand and use medicines better and provide practical treatment advice.
\end{enumerate}

The combination of medical LLMs and databases mentioned above can obtain information on drugs, dosages, contraindications, and side effects for treating corresponding diseases, which can be implemented in medical encyclopedias, assisted diagnosis, and other related tasks.

\paragraph{Medical Literature} As a high-quality medical data source, medical literature has become one of the most essential sources of knowledge extraction for researchers. It contains the latest findings and protocols that can supplement or update diagnostic and treatment principles in medical models. At the same time, the results of some medical experiments can also provide evidence to support the model reasoning in diagnosis and treatment recommendations. 

\begin{enumerate}
    \item \textbf{Medline}\footnote{https://www.nlm.nih.gov/medline/index.html} serves as the flagship bibliographic resource of the National Library of Medicine (NLM), featuring over 31 million citations for journal articles across the life sciences, with a particular emphasis on biomedical topics.
    \item \textbf{Embase}\footnote{https://www.embase.com/} is a scholarly repository created by Elsevier covering over 8,500 journals published in 95 countries and regions worldwide since 1974. The database contains over 32 million entries and over 2,900 proprietary journal indexes. The database includes 56,000 specialized search terms and 230,000 synonyms, greatly enhancing the accuracy and depth of searches. 
\end{enumerate}

\paragraph{Biomedical Information} KBs with knowledge of large quantities, small granularity, and high precision can help the medical model improve its mastery of medical knowledge. 

For example, the National Institutes of Health (NIH) has developed the \textbf{Unified Medical Language System}\footnote{https://www.nlm.nih.gov/research/umls/index.html} (UMLS) to unify various medical terminologies and provide standardized and precise information for medical research and applications. Its subsystems, including \textbf{MeSH}\footnote{https://www.nlm.nih.gov/mesh/} and \textbf{RxNorm}\footnote{http://www.nlm.nih.gov/research/umls/rxnorm/}, are widely used in medical literature retrieval, bioinformatics research, and EMR systems. 

The\textbf{ Human Phenotype Ontology}\footnote{https://hpo.jax.org} (HPO) was developed to aid researchers and clinicians in studying genetic diseases. It represents the complex changes in the human body during disease. HPO organizes phenotypic features into hierarchical structures, including morphological, physiological, and behavioral features. Using an object-oriented approach, researchers can communicate and share phenotypic information about genetic diseases more effectively, supporting disease diagnosis, treatment, and genetic counseling. 

To assist models with evidence-based reasoning of medicine, researchers developed \textbf{UpToDate Clinical Advisor}\footnote{https://www.uptodate.cn/home}. This is the Chinese version of UpToDate's product. It continuously combines existing medical evidence with the clinical experience of world experts. They present high-level, practical medical information that has undergone multiple rounds of filtering, digesting, and assimilating before being presented to users. In particular, the medical model can give graded diagnosis and treatment recommendations based on the GRADE principle of evidence-based medicine based on the comprehensive integration of research evidence based on the product.

\paragraph{Collective Database} Based on the above KBs, some comprehensive large-scale KBs were constructed to provide a complete view of medical knowledge and facilitate the model to learn the interactions and correlations between various types of knowledge, better serving real medical scenarios. 

For instance, \textbf{ClinicalKey}\footnote{https://www.clinicalkey.com/\#!/}, an online service platform developed by Elsevier, consolidates the most current medical information resources worldwide. Similarly, NHS Health, an online health information platform created by the NHS, offers information on diseases, treatments, and healthy lifestyles to assist the public in comprehending and maintaining their health. The information provided by NHS Health is based on clinical evidence. It is regularly reviewed and updated by a team of experts accredited by the UK Department of Health and Social Security.

The clinical KB is the cornerstone of medical LLMs, providing deep and reliable expertise and enabling continuous absorption and updating of medical knowledge. This dynamic learning process dramatically facilitates the application of medical LLMs in clinically assisted decision-making and personalized treatment recommendations, which is pivotal in providing high-quality healthcare services. 

\subsection{Clinical Datasets}
Building on the limited data available in each clinical database, medical LLMs often require a more comprehensive range of knowledge and information from rich and real medical data. Following the classification scheme for general LLM datasets, the datasets employed in constructing medical LLM are categorized into pre-training and fine-tuning datasets.

\subsubsection{Pre-training Datasets}
Pre-training of general LLMs refers to the extensive training of deep learning models on tongs of textual corpora to enable them to understand the generalized natural language expressions. Such pre-training gives the model a deep linguistic knowledge base, allowing it to perform better in subsequent tasks.

Similarly, medical LLMs require large amounts of corpora to learn vast medical knowledge and improve the domain's specific understanding of the model. The corpora can be obtained from medical-related literature, books, journals, websites,  and other related resources. Since textual content often does not fully cover fine-grained medical knowledge such as lesion features and organ images. Some pre-training datasets also combine multimodal information, such as aligned text-image data. Table \ref{tbl: pre_d} shows some commonly used pre-training datasets.

\begin{table*}[ht]
\footnotesize
\caption{Common pre-training datasets for medical LLMs. These datasets often contain vast textual information ("Text Corpora") or multimodal data, which is designed to provide a wide range of knowledge for medical LLMs during pre-training.}
\label{tbl: pre_d}
\centering
\begin{tabular}{cccc}
\hline
\textbf{Modality} & \textbf{Dataset} & \textbf{Data Scale} & \textbf{Typical Models}\\
\hline
\multirow{6}{*}{Text Corpora} & PubMed & $>$36M literatures & PubMedBERT \cite{PubMedBERT} \\
& MedDialog \cite{MedDialog} & 704.73M tokens & OphGLM \cite{OphGLM} \\
& EHRs \cite{EHRs} & $>$82B tokens & GatorTron \cite{EHRs} \\
& ChiMed-CPT \cite{ChiMed-CPT} & 3GB & Qilin-Med \cite{ChiMed-CPT} \\
& GAP-REPLAY \cite{MediTron} & 48.1B tokens & Meditron \cite{MediTron} \\
& The Pile \cite{ThePile} & $>$109.5 GiB & BioMedLM \cite{BioMedLM} \\
\hline
\multirow{7}{*}{Multimodal}& MTB \cite{Med-Flamingo} & 584M tokens & Med-Flamingo \cite{Med-Flamingo} \\
& PMC-OA \cite{PMC-CLIP} & 1.6M medical image-text & PMC-CLIP \cite{PMC-CLIP} \\
& MIMIC-CXR \cite{MIMIC-CXR} & $>$377k medical image-text & MedCLIP \cite{MedCLIP} \\
& MIMIC-III \cite{MIMIC-III} & $>$53.4k EHRs & ClinicalBERT \cite{ClinicalBERT} \\
& \makecell{MIMIC-IV \cite{MIMIC-IV} \\ MIMIC-CXR-JPG \cite{mimic-cxr-jpg} \\ MIMIC-IV-Note \cite{MIMIC-IV-Note}} & \makecell{$>$40k EHRs \\ $>$377k medical images \\ $>$2.65M medical texts}& LLaMA-Care \cite{LLaMA-Care} \\
\hline
\end{tabular}
\end{table*}

\paragraph{Text Corpora} These corpora are the primary source of information for LLMs, including medical and linguistic knowledge. Most medical pre-training datasets contain text from medical literature, authentic consultation dialogues, and clinical practice records. Techniques like anomaly cleaning, representation normalization, and privacy encryption often pre-process these data. We list some typical text pre-training datasets as follows:

\begin{itemize}
    \item \textbf{PubMed}\footnote{https://pubmed.ncbi.nlm.nih.gov/download/}: PubMed collects over 36 million citations and abstracts of biomedical literature from Medline. Despite medicine, it also encompasses data such as nursing and other health disciplines. While PubMed contains journal article citations, it does not provide full-text access but includes a link to the full text.
    \item \textbf{MedDialog}: MedDialog comprises two extensive medical conversation datasets, MedDialog-EN and MedDialog-CN. MedDialog-EN is an English dataset with 260k dialogues, 510k corpus, and 44.53M tokens, covering approximately 96 specialty categories. MedDialog-CN is a Chinese dataset comprising 3.4 million dialogues, 11.3 million corpus, and 660.2M tokens, covering 29 major and 172 sub-specialty categories. Both datasets include real-life communication scenarios between patients and healthcare professionals with different characteristics, providing the model with knowledge for generating more human-like responses.
    \item \textbf{EHRs}: The EHRs were retrieved from the UF Health Integrated Data Repository (IDR), which comprises 200 million clinical notes from 2011 to 2021 for over 2 million patients. These records come from over 126 clinical departments and approximately 50 million visits, covering healthcare settings such as inpatient, outpatient, and emergency room visits. In total, there are over 82 billion medical words. This dataset can assist models in understanding the correlation between data gathered in actual healthcare situations and the ultimate medical diagnosis.
    \item \textbf{ChiMed-CPT}: ChiMed-CPT is a unique dataset comprising multiple sub-datasets of four types: QA, Text, KG, and Dialogue. These sub-datasets can help medical models enhance multiple downstream tasks. The QA category includes Huatuo-26M-encyclopedias, Huatuo-26M-medical\_knowledge, and CMExam. The text category includes MedQA textbooks. The KG category includes CPubMed-KG, Xywy-KG, and 39Health-KG. The dialogue category includes Chinese-medical-dialogue-data, Medical-Dialogue-System, and CMD.
    \item \textbf{GAP-REPLAY}: The GAP-REPLAY dataset incorporates 48.1B tokens from four datasets: Clinical Guidelines, which includes 46K clinical practice guidelines from various healthcare-related sources, Paper Abstracts, comprising openly accessible abstracts from 16.1M closed-access PubMed and PubMed Central papers, Medical Papers, consisting of full-text articles from 5M publicly available PubMed and PubMed Central papers, and a Replay dataset including general domain data distilled to compose 1\% of the entire corpus.
\end{itemize}

In addition to the above datasets constructed specifically for medical scenarios, several relevant medical-themed data exist in general domain pre-trained datasets that can also be used to build medical LLMs.

One of the commonly used corpora for building general domain LLMs is \textbf{The Pile}. It is a pre-training dataset for general domain LLMs. It contains 825.18 GiB of English text, divided into 22 sub-datasets covering academic (such as PubMed Central, and ArXiv), internet (such as Pile-CC and OpenWebText2), prose (such as Books3 and BookCorpus2), dialogue (such as OpenSubtitles and YouTube Subtitles), and other types (such as GitHub, DeepMind Mathematics, and Enron Emails).

Notably, the \textbf{PubMed Central} and \textbf{PubMed Abstracts} datasets contain medical-related knowledge, which accounts for approximately 17.47\% or about 109.53 GiB. Models such as BioMedLM-2.7B are constructed using purely medical-themed datasets and achieve state-of-the-art (SOTA) results on the medical question-answering task.

\paragraph{Multimodal Data} Multimodal data can assist models in acquiring detailed knowledge of medical images and modeling the text-image relationships through pre-training. The data is often obtained from official sources such as textbooks and literature and real data sources such as radiology departments. To create these multimodal medical pre-training datasets, we must pre-process the images by denoising, resampling, enhancing, and normalizing them. Additionally, we should follow the textual data pre-processing steps mentioned earlier. Below are some brief descriptions of a few multimodal pre-training datasets.

\begin{itemize}
    \item \textbf{MTB}: A multimodal dataset Medical TextBooks (MTB) constructed from 4,721 textbooks from different medical specialties. During pre-processing, each book with cleaned text and images is segmented into several segments for pre-training so that each segment contains at least one image, thus matching the images with the associated text and unifying the semantic vectors of the text and images. The final MTB contains about 800,000 images and 584M tokens.
    \item \textbf{PMC-OA}: The PubMed Central Open Access (PMC-OA) dataset is derived from medical literature and contains 1.6M image-text pairs. This dataset is eight times larger than similar datasets and covers various diseases, organs, and imaging modalities. PMC-OA offers advantages in sample richness, number of modalities, disease types, and patient balance, regardless of diagnostic means, diseases, or patients' age and gender. The dataset provides detailed organ imaging features and reduces the problem of out-of-distribution generalisability for the model due to excessive data for some diseases.
    \item \textbf{MIMIC-CXR}: Medical Information Mart for Intensive Care Chest X-ray (MIMIC-CXR) is the largest chest X-ray dataset for modeling, providing comprehensive and accurate medical descriptions of chest X-ray features and corresponding symptoms, including chest X-rays in dicom format and free text radiological reports. The dataset comprises 377,110 image-text pairs, with each image being accompanied by a clinical report that describes the physician's findings.
    \item \textbf{MIMIC-III}: Medical Information Mart for Intensive Care (MIMIC-III) is an extensive, single-center database comprising information about patients admitted to critical care units at a large tertiary care hospital. Data includes vital signs, medications, laboratory measurements, observations and notes charted by care providers, fluid balance, procedure codes, diagnostic codes, imaging reports, hospital length of stay, survival data, and more. MIMIC-III includes data from more than 50,000 hospital admissions for adult patients (16 years old or above) admitted to critical care units between 2001 and 2012. Additionally, it includes data for over 7,870 neonates admitted between 2001 and 2008. The dataset can support broad research efforts, including epidemiology, clinical decision rule optimization, and medical e-tool development.
    \item \textbf{MIMIC-IV}: MIMIC-IV is an updated version of MIMIC-III, featuring current data and enhancements in various aspects. It comprises de-identified health data from more than 40,000 patients who are admitted to intensive care units at Beth Israel Deaconess Medical Center. The data includes detailed patient demographics, hospitalizations, lab measurements, and medication prescriptions.
    \item \textbf{MIMIC-CXR-JPG}: MIMIC-CXR-JPG consists of 377,110 chest radiographs in JPG format derived from the MIMIC-CXR dataset. This dataset aims to facilitate medical image understanding and analysis research by providing a more accessible JPG format of the original DICOM images. This dataset contains metadata and structured labels, all fully de-identified.
    \item \textbf{MIMIC-IV-Note}: MIMIC-IV-Note is a collection of de-identified free-text clinical notes for patients in the MIMIC-IV clinical database. It contains 331,794 de-identified discharge summaries from 145,915 patients admitted to the hospital and emergency department at the Beth Israel Deaconess Medical Center in Boston, MA, USA. Additionally, it also contains 2,321,355 de-identified radiology reports for 237,427 patients.
\end{itemize}

\subsubsection{Fine-tuning Datasets}
\label{finetuning_datasets}
Unlike the pre-training phase, which enables the model to learn a generic feature representation, the fine-tuning phase adapts the model with a suitable fine-tuned dataset to enhance its performance on a specific task. The fine-tuning datasets usually consist of well-labeled task-specific data, which are then adapted to the particular task by updating some of the parameters of the pre-trained model to improve the prediction accuracy of the model in the particular task. Table \ref{tbl: ft_d} demonstrates the fine-tuning datasets used in the medical for nine tasks: medical examination (ME), medical conversation (MC), medical question answering (MQA), medical visual question answering (MVQA), medical visual question generation (MVQG), image-to-text retrieval \& text-to-image retrieval (I2T \& T2I), report summarization (RS), report generation (RG), and medical image classification (MIC).

\begin{table}[h]
\footnotesize
\caption{Common fine-tuning datasets for medical LLMs. These datasets are often used to improve or evaluate the model performance on specific tasks and they often support multi-task analysis.}
\label{tbl: ft_d}
\centering
\resizebox{.85\textwidth}{32mm}{
\begin{tabular}{ccccccccccc}
\hline
\multirow{2}{*}{\textbf{Modality}} & \multirow{2}{*}{\textbf{Dataset}} & \multicolumn{9}{c}{\textbf{Task}}\\
& & \textbf{ME} & \textbf{MC} & \textbf{MQA} & \textbf{MVQA} & \textbf{MVQG} & \textbf{I2T \& T2I} & \textbf{RS} & \textbf{RG} & \textbf{MIC}\\
\hline
\multirow{8}{*}{Text Corpora} & CMExam \cite{CMExam} & \Checkmark &&&&&&&& \\
& MedQA \cite{MedQA} & \Checkmark &&&&&&&& \\
& PubMedQA \cite{PubMedQA} &&&\Checkmark&&&&&& \\
& cMedQA2 \cite{cMedQA2} &&&\Checkmark&&&&&& \\
& MedQuAD \cite{MedQuAD} &&&\Checkmark&&&&&& \\
& CMD. &&\Checkmark&\Checkmark&&&&&& \\
& MedDialog-CN \cite{MedDialog} &&\Checkmark&&&&&&& \\
& MultiMedQA \cite{Med-PaLM} &\Checkmark&&\Checkmark&&&&&& \\
\hline
\multirow{10}{*}{Multimodal} & PathVQA \cite{PathVQA} &&&&\Checkmark&&&&& \\
& VQA-RAD \cite{VQA-RAD} &&&&\Checkmark&&&&& \\
& VQA-med-2018 \cite{VQA-med-2018} &&&&\Checkmark&&&&& \\
& VQA-med-2019 \cite{VQA-med-2019} &&&&\Checkmark&&&&& \\
& VQA-med-2020 \cite{VQA-med-2020} &&&&\Checkmark&\Checkmark&&&& \\
& VQA-med-2021 \cite{VQA-med-2021} &&&&\Checkmark&\Checkmark&&&& \\
& SLAKE \cite{SLAKE} &&&&\Checkmark&&&&& \\
& PMC-15M \cite{PMC-15M} &&&&&&\Checkmark&&& \\
& ChiMed-VL \cite{ChiMed-VL} &&&&\Checkmark&&&&& \\
& MultiMedBench \cite{MultiMedBench} &&&\Checkmark&\Checkmark&&&\Checkmark&\Checkmark&\Checkmark \\
\hline
\end{tabular}
}
\end{table}

We categorize these fine-tuned datasets into five groups for various medical tasks: medical exams, medical question answering, medical conversations, medical visual question answering, and image-text retrieval.

\paragraph{Medical Exams} Medical exams are crucial for verifying the professional competence of medical students. They rely on real-world data, cover a broad range of knowledge, use standardized assessments (including standard answers and scoring criteria), provide constantly updated information resources, and involve multidisciplinary cross-fertilization, which has proven effective for evaluating medical LLMs. For example:

\begin{itemize}
    \item \textbf{CMExam}: The Chinese Medical Exam (CMExam) is a Chinese National Medical Licensing Examination dataset. The dataset contains over 60,000 multiple-choice questions and five additional annotations, including disease groups, clinical departments, medical disciplines, competency areas, and question difficulty levels. It is also the first Chinese medical exam dataset to provide comprehensive annotation by medical personnel, primarily for standardized and objective assessment and answer explanations for open model reasoning assessment.
    \item \textbf{MedQA}: The MedQA dataset was collected from professional medical board exams, including the United States Medical Licensing Examination (USMLE), the Mainland China Medical Licensing Examination (MCMLE), and the Taiwan Medical Licensing Examination (TWMLE). It consists of 61,097 multiple-choice questions covering three languages, with 12,723 in English, 34,251 in traditional Chinese, and 14,123 in simplified Chinese.
\end{itemize}

\paragraph{Medical Question Answering} Medical question answering, as one of the most common healthcare practices, often involves patients obtaining medical encyclopedia knowledge, disease differentiation, and medication recommendations. Researchers have designed the following fine-tuned datasets to enhance the model's performance in these scenarios.
\begin{itemize}
    \item \textbf{PubMedQA}: PubMedQA collects abstracts from PubMed literature. Models are required to answer questions based on the article abstracts. The dataset is divided into three subsets: 1k expert labeled, 61.2k unlabeled, and 211.3k artificially generated QA instances. This dataset aims to enhance the model's understanding of medical literature and improve the identification of key scientific issues through simple classification problems.
    \item \textbf{cMedQA2}: The cMedQA2 dataset is designed for routine medical scenarios and contains real drug-related questions from a Chinese medical search website\footnote{http://www.xywy.com/}. It comprises 108,000 questions and 203,569 answers, with qualified doctors providing the real answers to relevant questions after privacy treatment. Researchers can use this dataset to evaluate their models' medication knowledge and QA ability.
    \item \textbf{MedQuAD}: For more complex and specialized medical scenarios, the Medical Question Answering Dataset (MedQuAD) collates more than 40,000 medical QA pairs from the NIH's 12 websites (such as National Cancer Institute (NCI), Genetics Home Reference (GHR), Genetic and Rare Diseases Information Center (GARD), and MedlinePlus Health Topics). These data cover 37 question types (such as treatment, diagnosis, and side effects) related to diseases, medications, and other medical entities (such as examination).
\end{itemize}

\paragraph{Medical Conversations} Medical conversations are the primary means of daily treatment and typically involve multiple QA pairs with larger contexts. Doctors use them to gather information about patients' conditions and provide medical advice. Researchers have constructed the following medical dialogue dataset to enhance the ability of intelligent treatment and assisted decision-making.

\begin{itemize}
    \item \textbf{CMD.}\footnote{https://github.com/Toyhom/Chinese-medical-dialogue-data}: The Chinese medical dialogue dataset comprises 792,099 data pieces that cover six types of medical dialogues: Andriatria (90,000 QA pairs), Internal Medicine (220,000 QA pairs), Obstetrics and Gynecology Department (180,000 QA pairs), Oncology (70,000 QA pairs), Pediatric (100,000 QA pairs), and Surgical (110,000 QA pairs). This dataset can be used to study the model's expertise and KB of the questioning process across a wide range of departments.
    \item \textbf{MedDialog-CN}: To address the issue of varying linguistic expressions and causes of morbidity among individuals, MedDialog-CN is a database that collects patient data from diverse demographics, including age, gender, occupation, education, and income, across 31 provincial-level administrative regions in China. This database is used to construct Chinese medical dialogue datasets containing 1.1 million dialogues and 4 million utterances, covering 29 major specialties and 172 sub-specialties.
\end{itemize}

\paragraph{Medical Visual Question Answering} Medical images often contain numerous patient situations and fine-grained disease features. They are a crucial component of medical diagnosis. Therefore, researchers have developed multiple multimodal medical QA datasets to enhance the extraction and alignment of multimodal information for the model.

\begin{itemize}
        \item \textbf{PathVQA}: PathVQA is a pathology VQA dataset that generates QA pairs from captions extracted from pathology textbooks and online digital libraries using NLP. The dataset contains 4,998 pathology images and 32,799 QA pairs, with each question manually checked for correctness. This dataset provides the model with rich disease image-text pairing data, improving the model's modal alignment.
        \item \textbf{VQA-RAD}: VQA-RAD is a unique dataset where clinicians ask natural questions about radiological images and provide reference answers. The dataset includes a human transverse image set containing samples of the head, chest, and abdomen from MedPix\footnote{https://medpix.nlm.nih.gov/}. The final VQA-RAD dataset comprises 315 images and 3,515 question pairs.
        \item \textbf{VQA-med series}: ImageCLEF released datasets for real medical scenarios during the 2018, 2019, 2020, and 2021 Challenges: VQA-med-2018, VQA-med-2019, VQAmed-2020, and VQA-med-2021. The datasets have been updated over time, with increasing volumes of data and more refined medical theories. For instance, VQA-med-2019 is based on VQA-RAD and focuses on four prevalent problem categories: modality, plane, organ system, and abnormality. The first three categories can be addressed as a classification task, while the fourth category requires generating answers. Moreover, the dataset's task complexity has increased with the inclusion of the VQG task in VQA-med-2020. The VQA-Med-2021 dataset has been enhanced with a validation set and a test set, following the principles of VQA-Med-2020. A medical professional manually reviewed these sets, resulting in improved accuracy and task complexity.
        \item \textbf{SLAKE}: SLKAE is a bilingual dataset with semantic labels annotated by experienced physicians and new structured medical knowledge. It covers more body parts and richer modalities than other datasets. The dataset consists of 642 images and 14,028 questions related to 12 diseases and 39 organs and contains 2,603 English triples and 2,629 Chinese triples.
\end{itemize}

\paragraph{Image-text Retrieval} Image-text retrieval in traditional multimodal medical modeling involves detecting a model's multimodal alignment. This requires the model to retrieve the most matching text or image data from a large dataset based on a particular image or text. To achieve this, the researchers developed the PMC-15M dataset to fine-tune the model.

\textbf{The PMC-15M dataset} was created using PubMed Central. This involved downloading and extracting a compressed catalog containing full article packages. Each package contains XML, PDF, media, and supplementary material. The dataset also includes image files with corresponding captions and the PMIDs and PMCIDs of the original articles. The PMC-15M dataset comprises 15 million image-caption pairs for over 3 million articles.

Furthermore, various extensive datasets are frequently utilized in developing medical models. These datasets are crucial for enhancing the model's generalizability and performance across multiple tasks.

\begin{itemize}
    \item \textbf{MultiMedQA}: MultiMedQA is a textual dataset comprising seven medical QA datasets, including six existing datasets and one newly introduced dataset: MedQA, MedMCQA \cite{MedMCQA}, PubMedQA, LiveQA \cite{LiveQA}, MedicationQA \cite{MedicationQA}, MMLU \cite{MMLU}, and HealthSearchQA. HealthSearchQA contains 3,375 commonly searched medical questions by users. This dataset was used to evaluate the diversity of LLM's clinical knowledge and QA abilities. The emphasis was on question complexity and response length during data construction, resulting in approximately 213k data points.
    \item \textbf{ChiMed-VL}: ChiMed-VL was created by translating several open-source English medical multimodal datasets into Chinese using GPT-3.5 with expert quality control. The dataset comprises two parts: ChiMed-VL-Alignment and ChiMed-VL-Instruction. The data for ChiMed-VL-Alignment is derived from PMC-OA and PMC-CaseReport and contains 580,000 image-text pairs. Each pair includes contextual information or a description of the image, providing detailed alignment data for the model. The ChiMed-VL-Instruction dataset is sourced from PMC-Report and PMC-VQA and comprises 460,000 QA pairs. These datasets encompass various diagnostic modalities, including X-rays, MRIs, CTs, radioisotopes, mitosis, and other biomedical knowledge.
    \item \textbf{MultiMedBench}: MultiMedBench is a benchmark created to develop and evaluate general biomedical AI. The benchmark is multimodal and multi-task, comprising 12 de-identified open-source datasets for 5 task types, including QA, RS, VQA, RG, and MIC, with 14 individual tasks. It assesses the capacity of medical LLMs to carry out a range of clinically relevant tasks. The benchmark includes over 1 million samples covering various topics, including medical issues, radiology reports, pathology, dermatology, chest X-rays, mammography, and genomics.
\end{itemize}

\subsubsection{Findings}
We have summarized and analyzed the above clinical datasets, including both the pre-training and fine-tuning datasets, in terms of data type, authenticity, diversity, coverage, update frequency, usability, reliability, and language, as shown in Table \ref{tab:summary}. Additionally, we have presented examples of these datasets in github\footnote{https://github.com/vicky-yuan/survey-datasets} to provide further insight into their format and applications.

As can be seen in Table \ref{tab:summary}, the available datasets exhibit the following characteristics:
\begin{itemize}
    \item \textbf{Data Sources and Types.} Most datasets are based on real data and primarily contain textual information, while image data is mainly concentrated in radiology.
    \item \textbf{Coverage.} The coverage of the datasets can be divided into two categories: one encompasses a broad range of medical foundational knowledge, while the other focuses on specific specialty knowledge.
    \item \textbf{Time Span and Update Frequency.} The datasets vary widely in their timeframe and are generally updated less frequently, except for PubMed, which is kept steadily updated, and other datasets with irregular update cycles.
    \item \textbf{Accessibility.} Most datasets are open-source and available, while a few require permission to access.
    \item \textbf{Data Reliability.} Most datasets are considered reliable, as they are based on real data. However, the reliability of a few datasets is partially reliable due to some data being generated by models like GPT-3.5/GPT-4.
    \item \textbf{Language Distribution.} The datasets are predominantly in English, followed by bilingual (English and Chinese), while datasets in other languages are relatively scarce.
\end{itemize}

\begin{sidewaystable}[htbp]
\tiny
\begin{threeparttable}
\caption{A summary of the above clinical datasets, which include both pre-training datasets (text pre-training datasets and multimodal pre-training datasets) and fine-tuning datasets (text fine-tuning datasets and multimodal fine-tuning datasets), in terms of data type, authenticity, diversity, coverage, update frequency, usability, reliability, and language. ``Y'' means ``yes'', ``N'' means ``no'', ``M'' means ``mixed datasets (i.e., both real and generated data)'', ``CA'' means ``conditional access'', and ``P'' means ``partially reliable''.}
\label{tab:summary}
\centering
\begin{tabular}{cccccccccccccccc}
\hline
\multirow{2}{*}{} & \multirow{2}{*}{\textbf{DataType}} & \multirow{2}{*}{\textbf{Authenticity}} & \multicolumn{6}{c}{\textbf{Diversity}} & \multicolumn{3}{c}{\textbf{Medical Coverage}} & \multirow{2}{*}{\textbf{Update Frequency}} & \multirow{2}{*}{\textbf{Usability}} & \multirow{2}{*}{\textbf{Reliability}} & \multirow{2}{*}{\textbf{Language}} \\
& & & Text & Figure & Radiology & Pathology & Time series & Numerical Value & Fundamentals & Specialty Knowledge & Time Span & & & & \\
\hline
PubMed & Medical Literature & Y & Y & - & - & - & - & Y & Y & - & - & \makecell{Once a year, \\ around December} & Y & Y & English \\
MedDialog & Doctor-Patient Dialogue & Y & Y & - & - & - & - & - & - & 96/172 diseases & \makecell{2008-2020/\\2010-2020} & - & Y & Y & English/Chinese \\
EHRs &  Patient Clinical Records & Y &  Y & - & - & - & - & - & - & 126 clinical departments & 2011-2021 & - & N & Y & English \\
ChiMed-CPT & Combined Dataset & Y &  Y & - & - & - & - & - & Y & - & - & - & Y & Y & English/Chinese \\
GAP-REPLAY & Combined Dataset & Y &  Y & - & - & - & - & Y & Y & - & - & - & Y & Y & English \\
The Pile & Combined Dataset & Y &  Y & - & - & - & - & Y & Y & - & - & - & Y & Y & English \\
\hline
MTB & Medical Textbooks & Y & Y & Y & - & - & - & - & Y & - & - & - & N & Y & English \\
PMC-OA & Literature Image-Text & Y & Y & Y & - & - & - & - & Y & - & - & - & Y & Y & English \\
MIMIC-CXR &  Chest X-ray Image-Text & Y &  Y & - & Y & - & - & - & - & 14 types of chest findings & 2011-2016 & - & CA & Y & English \\
MIMIC-III & Patient Health Data & Y &  Y & - & Y & - & Y & Y & Y & - & 2001-2012 & - & CA & Y & English \\
MIMIC-IV & Patient Health Data & Y & Y & - & Y & - & Y & Y & Y & - & 2008-2019 & - & CA & Y & English \\
MIMIC-CXR-JPG & Chest X-ray Images & Y & - & - & Y & - & - & - & - & 14 types of chest findings & 2011-2016 & - & CA & Y & English \\
MIMIC-IV-Note & Patient Clinical Records & Y & Y & - & Y & - & Y & Y & Y & - & \makecell{Within one year \\ after the visit} & - & CA & Y & English \\
\hline
CMExam & Medical Licensing Exam QA & Y & Y & - & - & - & - & - & - & \makecell{26 diseases, \\ 35 clinical departments} & - & - & Y & Y & Chinese \\
MedQA & Medical Board Exam QA & Y & Y & - & - & - & - & - & Y & - & - & - & Y & Y & \makecell{Simplified Chinese/\\Traditional Chinese/\\English} \\
PubMedQA &  Medical Literature QA & M & Y & - & - & - & - & Y & Y & - & - & - & Y & P & English \\
cMedQA2 & Chinese Community Medicine QA & Y & Y & - & - & - & - & - & Y & - & - & - & Y & Y & Chinese \\
MedQuAD & Medical QA Pairs & Y & Y & - & - & - & - & - & Y & - & - & - & Y & Y & English \\
CMD. & Medical Dialogue & Y & Y & - & - & - & - & - & - & 6 types of clinical departments & - & - & Y & Y & Chinese \\
MedDialog-CN & Doctor-Patient Dialogue & Y & Y & - & - & - & - & - & - & \makecell{29 broad categories of specialties \\ \& 172 fine-grained specialties} & 2010-2020 & - & Y & Y & Chinese \\
MultiMedQA & Combined Dataset & Y & Y & - & - & - & - & - & Y & - & - & - & Y & Y & English \\
\hline
PathVQA & Pathology Image QA & Y & Y & - & - & Y & - & - & - & Pathology & - & - & Y & Y & English \\
VQA-RAD & Radiology QA & Y & Y & - & Y & - & - & - & - & Radiology & - & - & Y & Y & English \\
VQA-med-2018 & Medical Image QA & Y & Y & - & Y & - & - & - & - & Radiology & - & - & Y & Y & English \\
VQA-med-2019 & Medical Image QA & Y & Y & - & Y & - & - & - & - & Radiology & - & - & Y & Y & English \\
VQA-med-2020 & Medical Image QA & Y & Y & - & Y & - & - & - & - & Radiology & - & - & Y & Y & English \\
VQA-med-2021 & Medical Image QA & Y & Y & - & Y & - & - & - & - & Radiology & - & - & Y & Y & English \\
SLAKE & Medical Image QA & Y & Y & - & Y & - & - & - & - & 12 diseases, 39 organs & - & - & Y & Y & English/Chinese \\
PMC-15M & Biomedical Image-Text Pairs & Y & Y & Y & - & - & - & - & Y & - & - & - & N & Y & English \\
ChiMed-VL & Combined Dataset & Y & Y & Y & Y & - & - & - & Y & - & - & - & Y & P & English/Chinese \\
MultiMedBench & Combined Dataset & Y & Y & - & Y & Y & - & - & Y & \makecell{Pathology, radiology, \\ dermatology, genomics} & - & - & CA & Y & English \\
\hline
\end{tabular}

\end{threeparttable}
\end{sidewaystable}

\subsection{Summary}
Besides utilizing the above training data, the instruction fine-tuning phase also involves utilizing instructional datasets to enhance the model's generalization among tasks. The general approach is to automatically generate instruction data using pre-trained models such as Self-Instruct framework \cite{Self-Instruct}, BELLE \cite{BELLE}, or GPT \cite{GPT}. These models are used to model the responses of real users for various tasks, resulting in instruction datasets that can understand and execute human instructions. For instance, ChatDoctor \cite{ChatDoctor} has created prompt templates using the fine-tuned LLaMA-7B \cite{LLaMA-7B}, which incorporates the alpaca instruction dataset and the HealthCareMagic100k doctor-patient dialogue dataset. These templates are designed to retrieve external knowledge databases during doctor-patient dialogues, resulting in more accurate model outputs. Huatuo \cite{Huatuo} utilized the LLaMA-7B to generate a dataset for Chinese medical instructions. The model was fine-tuned with the Medical KG and GPT3.5 API and subsequently enhanced to improve the QA effectiveness of LLaMA in the medical field.

However, current medical KBs and training datasets do not yet have sufficient multilingual support. For example, most existing resources are only available in English or Chinese, with few aligned datasets. This limitation can hinder the applications of medical AI systems for non-native doctors and patients. Furthermore, the integrity of disease knowledge in these KBs and datasets may not be sufficient enough. For instance, information on certain rare or endemic diseases, such as Gaucher disease, Pompe disease, and Fabry disease, is insufficient for model learning, and updating the latest research results and treatments is also costly. In real healthcare scenarios, these shortcomings may impact diagnostic accuracy, treatment effectiveness, and healthcare quality. We will discuss these limitations in Chapter \ref{chp: 6}.

\section{Academic Medical LLMs}
\label{chp: 3}
LLM refers to deep neural networks based on the Transformer architecture \cite{Vaswani17Attention}, designed to process and generate natural language text. Unlike traditional language models, LLMs are trained using large-scale textual data with parameter sizes up to billions. For instance, models like GPT-3 \cite{GPT-3}, GLM \cite{GLM}, LLaMA, and BLOOM \cite{BLOOM} are trained on large corpora such as Wikipedia\footnote{https://huggingface.co/datasets/wikipedia} and BookCorpus \cite{BookCorpus}. Moreover, recent studies \cite{ChatGPT} have demonstrated that these models can exhibit exceptional generalizability across multiple domains through pre-training and fine-tuning.

Similarly, LLMs also have attracted extensive research \cite{He23ASurvey} \cite{EHRs} in the healthcare field, and these models can process rich medical literature, clinical records, and other health-related data, providing powerful support for medical research, diagnosis, and treatment. Medical LLMs can help medical professionals understand patient conditions and make decisions after integrating clinical knowledge. Below, we will introduce and analyze the mechanisms of building medical LLM from two perspectives: text corpora and multimodal data.

\subsection{Mechanisms for Building Medical LLMs}
In the previous section, we examined open-source medical pre-training datasets and fine-tuning datasets to enhance the clinical knowledge of LLM. Based on our analysis of existing studies, to achieve optimal results, it is necessary to select a suitable LLM backbone in the general domain or a model with excellent performance in a specific domain at the beginning. Then, the backbone model should be trained for specific medical goals. For example, enhancing the clinical knowledge of the model requires mixing clinical data corpora with the pre-training dataset that trains the model backbone for the supplement, which can significantly improve the model's understanding of clinical applications with minimal computation cost.

To better illustrate the building mechanisms for LLM from clinical knowledge, we classify this process into two categories based on the representation of clinical knowledge: text corpora and multimodal. These two building mechanisms also require customized training objectives for different types of clinical data and application scenarios, enabling the model to perform excellently in specific clinical tasks.

\subsubsection{Building Medical LLMs from Text Corpora}
Text corpora play a crucial role in medical research and building domain LLMs. This data provides an in-depth analysis and summary of multiple findings and knowledge in medical research. In the following section, we first present the training objectives during pre-training with standard practices. We then present the fine-tuning strategies for various downstream medical tasks. Finally, we introduce several comprehensive works that combine pre-training and fine-tuning into a multi-stage training pipeline to obtain more capable medical LLMs.

\paragraph{Pre-training}
BioMedLM \cite{BioMedLM} and GatorTronGPT \cite{GatorTronGPT} are the main studies using these pre-training methods to build medical LLMs with strong multi-task generalization capabilities.

BioMedLM, previously named PubMedGPT, is a biomedical language model developed jointly by the Stanford Center for Basic Modeling Research and MosaicML. It is based on GPT-2 \cite{GPT-2} and pre-trained using the PubMed Abstracts and PubMed Central portions of the Pile dataset, which contains approximately 50 billion tokens covering 16 million abstracts and 5 million full-text articles from the biomedical literature. The final model achieves an accuracy of 50.3\% on the MedQA-USMLE, 74.4\% on the PubMedQA, and 95.7\% on the BioASQ. These results demonstrate the capability of LLM, especially in the biomedical domain, and their language generation capabilities in real-world applications.

GatorTronGPT was trained as a generative clinical LLM on 277 billion words of textual data. This included 82 billion words of clinical texts from 126 clinical departments and about 2 million patients at the medical center and 195 billion words of various general English texts. Their evaluation of the training results based on the GPT-3 in biomedical NLP and medical text generation showed that GatorTronGPT was comparable to humans regarding linguistic readability and clinical relevance. Physicians were unable to distinguish between them.

\paragraph{Fine-tuning} In addition to pre-training, fine-tuning methods are commonly used to construct medical LLMs for specific downstream tasks. Table \ref{tbl:train_task} below, these tasks include information extraction (IE), MQA, multi-turn dialogue (MD), ME, and text-to-text (T2T).

\begin{table*}[!htb]
\footnotesize
	\caption{Training tasks for various medical tasks. Each of the tasks uses different training objectives and serves various purposes in real-life applications.}
    \label{tbl:train_task}
	\centering
	\begin{tabularx}{\textwidth}{c c >{\centering\arraybackslash}X}
        \hline
		\textbf{Task} & \textbf{Training Objective} & \textbf{Medical Application Scenarios}  \\
        \hline
		IE & \makecell[c]{Sequence Labelling \\ Generative Extraction} & Case Structurization \\

            MQA & Text Generation &\makecell{Knowledge QA, Assisted Diagnostic Suggestions, \\ Drug Indication Evaluation, Disease Evaluation, Report Interpretation} \\

            MD & Long Context Text Generation & Simulated Diagnosis and Treatment, Guidance \\

            ME & Text Classification & Learning Assistance \\

            T2T & Text Generation & Synthesizing Clinical Text Generation \\
        \hline
	\end{tabularx}%
\end{table*}%

\textbf{Information Extraction (IE)} automatically extracts structured information from unstructured or semi-structured documents and other electronically represented sources. Medical IE tasks can enhance the quality of models for applications such as case description structurization and medical KB construction.

The conventional approach to IE involves training the model using sequential labeling. The researcher uses BILOU to tag important information in a given text and then models the tagging patterns to extract the tagged text as the final result. BiLSTM+CRF \cite{BiLSTM+CRF} techniques are frequently employed for this type of IE.

Following the emergence of generative modeling, researchers began using generative extraction methods to gain more direct and efficient access to extracted information. Generative extraction takes the input text and serializes the structured result of the extraction as the final output. SOTA generative extraction approaches \cite{SOTA} typically use a method similar to the UIE \cite{UIE} architecture.

\textbf{Medical Question \& Answering (MQA)} involves generating a response to a question based on the model's given context or knowledge stock. This task can enhance the performance of models in scenarios such as medical encyclopedias, assisted diagnosis, drug indications, disease assessment, and report interpretation.

QA tasks with context can also be used to train models using the sequence annotation method described above. However, researchers often prefer to use generative methods (such as Sunsimiao\footnote{https://github.com/X-D-Lab/Sunsimiao}, and QiZhenGPT\footnote{https://github.com/CMKRG/QiZhenGPT}) in conjunction with context-free QA tasks to improve training efficiency and quality.

\textbf{Multi-turn Dialogue (MD)} is an extension of the medical QA task, where multi-turn dialogues consist of multiple coherent QA pairs. The historical information of the dialogues is aggregated into contextual information and sent to the model as a new input for processing. This task requires a longer context length limit for the model, making it more challenging than the medical QA task. However, this task can greatly enhance the model's performance in complex medical scenarios, such as simulated consultation sessions, which is of greater practical value.

\textbf{Medical Examination (ME)} requires providing medical exam content as multiple choice questions to a model for analysis and selecting the best answer from the options. Traditional models typically split the question stem and option content into separate inputs for a sequential multi-classification task to determine the most suitable answer. The task involves testing the model's understanding of specific medical knowledge.

\textbf{Text-to-Text (T2T)} is the combination of text generation tasks other than tasks in the QA form. It requires the model to process text based on the given instruction. For example, for a given medical report and summarization instruction, the model should generate an abstract of the report. The instructions for textual tasks are usually expressed in declarative sentences. Many textual tasks do not require specifying a task instruction. Instead, the model can learn the requirements of a single instruction through training. These tasks are similar to the instruction fine-tuning tasks of the LLMs. Therefore, they are less computationally expensive to learn.

Most medical LLMs obtained through fine-tuning are designed for MQA and MD tasks, which are closely linked to real medical scenarios and, therefore, receive more attention. Table \ref{tbl:ft_llm} summarizes the construction of medical LLMs for specific downstream tasks using fine-tuning methods, which will be briefly presented as follows:

\begin{table*}[!h]
	\caption{Medical LLMs constructed using fine-tuning methods. These models often use an LLM backbone from the general domain as the foundation and are fine-tuned for different downstream medical tasks.}
    \label{tbl:ft_llm}
	\centering
  \resizebox{.7\textwidth}{50mm}{
	\begin{tabularx}{\textwidth}{c >{\centering\arraybackslash}X >{\centering\arraybackslash}X c}
        \hline
		\textbf{Model} & \textbf{Anchor Model} & \textbf{Data Source} & \textbf{Task} \\
        \hline
		Sunsimiao & \makecell{Baichuan-7B\\ChatGLM-6B\\InternLM-Chat-7B} & - & QA\\
        \hline
            QiZhenGPT & \makecell{Chinese-LLaMA-7B\\CaMA-13B\\ChatGLM-6B} & Qizhen Medical KB & QA\\
        \hline
            PMC-LLaMA \cite{PMC-LLaMA} & LLaMA-7B & \makecell{Books+Literature\\MedC-I \cite{PMC-LLaMA}} & QA \\
        \hline
            ChatMed-Consult & LLaMA-7B & ChatMed\_Consult\_Dataset & QA \\
        \hline
            BenTsao & LLaMA-7B & CMeKG-8K & QA \\
        \hline
            Med-PaLM & PaLM \cite{PaLM} & MultiMedQA & QA \\
        \hline
            Med-PaLM 2 \cite{Med-PaLM2} & PaLM2 \cite{PaLM2} & MultiMedQA & QA \\
        \hline
            ClinicalGPT \cite{ClinicalGPT} & Bloom-7B & \makecell{cMedQA2\\cMedQA-KG\\MD-HER\\MEDQA-MCMLE\\ MedDialog} & MD, QA \\
        \hline
            MedicalGPT & \makecell{Bloom\\LLaMA\\ChatGLM2-6B\\Baichuan-7B/13B} & \makecell{Chinese medical datasets\\HuatuoGPT-Hybrid SFT} & MD\\
        \hline
            HuatuoGPT \cite{HuatuoGPT} & \makecell{Baichuan-7B\\LLaMA-13B} & Hybrid SFT \cite{HuatuoGPT} & MD \\
        \hline
            BianQue \cite{BianQue} & ChatGLM-6B & BianQueCorpus \cite{BianQue} & MD \\
        \hline
            DoctorGLM \cite{DoctorGLM} & ChatGLM-6B & CMD. & MD \\
        \hline
            ChatDoctor \cite{ChatDoctor} & LLaMA & \makecell{HealthCareMagic\\iCliniq} & MD \\
        \hline
            PULSE & OpenChina LLaMA 13B & - & QA, MT, T2T \\
        \hline
\end{tabularx}%
}
\end{table*}%

\textbf{Sunsimiao:} Yan et al. developed the Sunsimiao Chinese Medical LLM to synthesize folk medical experiences and accumulate Chinese medical data. The model aims to provide a safe, reliable, inclusive Chinese medical LLM. It was mainly fine-tuned using Baichuan-7B\footnote{https://github.com/baichuan-inc/baichuan-7B}, ChatGLM-6B\footnote{https://github.com/THUDM/ChatGLM-6B}, and InternLM-Chat-7B\footnote{https://huggingface.co/internlm/internlm-chat-7b} on 100,000 high-quality Chinese medical data. However, the data is not open source.

\textbf{QiZhenGPT:} QiZhenGPT is an LLM constructed by Zhejiang University based on Chinese-LLaMA-Plus-7B\footnote{https://github.com/ymcui/Chinese-LLaMA-Alpaca}, CaMA-13B\footnote{https://github.com/zjunlp/CaMA}, and ChatGLM-6B models. It was fine-tuned using instructions from the Chinese medical instruction dataset constructed by the QiZhen Medical KB\footnote{http://www.mk-base.com/\#/official/home}. Unlike many open-source ChatLLM projects that use instruction data generated by other models, such as ChatGPT, this model utilizes real doctor-patient knowledge QA datasets from the Qizhen Medical KB to enhance its accuracy in Chinese medical scenarios.

\textbf{PMC-LLaMA:} PMC-LLaMA is a medical LLM released by Shanghai Jiao Tong University. It is based on the LLaMA-7B model and has been fine-tuned using 4.8 million biomedical academic papers. The model has been evaluated on three bioQA datasets: PubMedQA, MedMCQA, and USMLE. The results indicate a greater comprehension of biomedical domain-specific concepts, leading to a high level of performance in the QA benchmark. However, due to equipment performance, the limited training data and epochs for PMC-LLaMA also highlight the challenge of high costs and hardware requirements for fine-tuning LLMs.

\textbf{ChatMed:} East China Normal University has launched the ChatMed series of LLMs for Chinese medical to promote the development and landing of LLMs in the Chinese medical field. The ChatMed-Consult\footnote{https://github.com/michael-wzhu/ChatMed} is based on the ChatMed\_Consult\_dataset\footnote{https://huggingface.co/datasets/michaelwzhu/ChatMed\_Consult\_Dataset}, which contains over 500,000 online consultations and ChatGPT responses, to improve LLM's medical knowledge and ability to answer medical consultations. The model backbone used is LLaMA-7b, which combines the LoRA weights of Chinese-LLaMA-Alpaca with the Chinese extended word list. Efficient fine-tuning of parameters based on LoRA is then performed. Finally, after comparing with the Chinese-LLaMA-7B model on the effect of the questioning dialogue, it is evident that the ChatMed-Consult model is more effective in comprehending the user's query despite interference. Additionally, the responses are more humane and cautious, providing more practical suggestions.

\textbf{BenTsao:} BenTsao is a medical LLM developed by the SCIR Lab of Harbin Institute of Technology, based on the LLaMA-7B model. The model has been fine-tuned with Chinese medical instructions. They constructed a Chinese medical instruction dataset through medical KG and GPT3.5 API and further fine-tuned the model on this basis to improve the QA effect of LLaMA in the medical field.

\textbf{Med-PaLM series:} Google created Med-PaLM, a version of PaLM for fine-tuning in the medical field. It was the first model to achieve a passing score ($>60\%$) on US medical licensing-style questions. This study presented an evaluation dataset MultiMedQA, demonstrating the importance of a comprehensive benchmark for MQA. The dataset covers diverse MQA benchmarks for medical examination, consumer health, and medical research. The study also emphasized the significance of manual evaluation of model answers and alignment strategies in the medical domain. Med-PaLM accurately answered both multiple-choice and open-ended questions and explained its responses. However, the quality of the modeled answers still has some shortcomings when compared to those of physicians.
    
To address these gaps, Google has released PaLM2, a next-generation AI language model. Additionally, they have developed Med-PaLM 2, a medical domain variant based on PaLM2. To improve the reasoning ability of LLM, a new cueing strategy called "Ensemble refinement" has been proposed. Med-PaLM 2 consistently performed at the "expert" physician level on medical exam questions, scoring 85\%. This represents an 18\% improvement over Med-PaLM's previous performance and significantly outperforms similar AI models. The final results demonstrate that Med-PaLM 2 performs considerably better than Med-PaLM on both the multiple-choice and long-form assessments.

\textbf{ClinicalGPT:} ClinicalGPT, proposed by Wang et al., is a language model specifically designed and optimized for clinical scenarios. The model can effectively handle a wide range of clinical tasks by incorporating a diverse range of real-world data, including medical records, domain-specific knowledge, and multi-turn dialogue consultations during the training process. The model is based on BLOOM-7B and incorporates LoRA fine-tuning for training. Experimental results demonstrate its suitability for healthcare encyclopedic tasks.

\textbf{MedicalGPT}\footnote{https://github.com/shibing624/MedicalGPT}: Xv's MedicalGPT implements the training of language models for the medical industry using the ChatGPT Training Pipeline. The model is based on Bloom, LLaMA, ChatGLM\footnote{https://github.com/thudm/chatglm2-6b}, Baichuan 7B/13B\footnote{https://github.com/baichuan-inc/Baichuan-13B}, and other general-purpose LLMs. These models were fine-tuned using the ChatGPT training pipeline, which includes secondary pre-training, supervised fine-tuning, reward modeling, and reinforcement learning training. The training dataset comprises both medical and general-purpose datasets. The medical dataset consists of 2.4 million Chinese medical datasets\footnote{https://huggingface.co/datasets/shibing624/medical}, including pre-training, instruction fine-tuning, and reward datasets, as well as 220,000 Chinese medical dialogue datasets from the HuaTuo project. The results of the experiments indicate that MedicalGPT is significantly more effective in both daily QA and MQA. However, there are still issues with factual errors and unclear identification of hazardous commands, and the performance in scenarios involving reasoning and multi-turn dialogues needs improvement.

\textbf{HuatuoGPT:} Zhang et al. developed HuatuoGPT to provide the language model with the ability to diagnose and provide medical consultation advice in Chinese, similar to that of a physician. The model is constructed using Baichuan-7B and LLaMA-13B and fine-tuned with four types of corpora. These corpora consist of distilled data generated by ChatGPT, instruction, and dialogue data from real-world physicians responding to patients' questions. The aim is to maintain knowledge-rich communication with the user. Finally, following mutual validation of the automatic and manual evaluations, the models demonstrate strong performance in both single-turn QA and multi-turn interactive diagnostic scenarios.

\textbf{BianQue:} Bianque is a large-scale healthcare dialogue model initialized with ChatGLM-6B and fine-tuned with instructions from the BianQue corpus. BianQue-2.0 has expanded the instruction data with drug, medical encyclopedic knowledge, and ChatGPT distillation instructions, which improves the model's ability to make suggestions and answer knowledge queries. Unlike most language models, this model can be more closely related to everyday life and improve questioning skills using question chains.

\textbf{DoctorGLM:} DoctorGLM is the first Chinese diagnostic language model developed by the Shanghai University of Science and Technology. It is based on ChatGLM-6B and uses real-world online diagnostic dialogue data, including CMD, MedDialog, ChatDoctor, and HealthcareMagic. The model uses two different parameter fine-tuning methods, p-tuning and LoRA, to enable medical communication in natural dialogue. However, the diagnostic capability may need improvement.

\textbf{ChatDoctor:} ChatDoctor is an advanced language model designed specifically for medical applications, aiming to provide patients with an intelligent and reliable medical companion that offers personalized medical advice while answering medical queries. Based on the LLaMA model, it is trained and fine-tuned using over 110,000 real doctor-patient dialogue datasets (including the HealthCareMagic-100k dataset from the medical consultation website HealthCareMagic, the iCLINIQ-10K dataset from the medical consultation website iCliniq, and the GenMedGPT-5k dataset generated using ChatGPT from a disease database). It is equipped with an external knowledge base, which includes a knowledge library containing over 700 diseases, to accurately retrieve the corresponding knowledge and reliable sources to answer patient inquiries.

\textbf{PULSE}\footnote{https://github.com/openmedlab/PULSE}: The PULSE model is developed on the OpenMEDLab platform and is based on the OpenChina LLaMA 13B. The model is further fine-tuned using approximately 4 million SFT data from medical and general fields to support various NLP tasks in the medical field, including health education, doctor examination questions, report interpretation, medical record structure, and simulated diagnosis and treatment. However, due to the relatively small model size, although the model provides reasoning results about disease diagnosis and treatment, these results cannot replace offline professional doctors' advice and treatment plans. All responses are for reference only and should not be used as the basis for diagnosis or treatment.

\paragraph{Multi-stage Training} In addition to the methods mentioned above, some researchers construct more comprehensive knowledge-gaining and widely applicable medical models across tasks by using a combination of pre-training and fine-tuning in multi-stage training, as shown in Table \ref{tbl:preft_llm}. We will describe them briefly:

\begin{table}[h]
\footnotesize
    \centering
    \caption{Pre-training and fine-tuning integrated for Medical LLMs. These models often integrate clinical data in multiple training stages to ensure the accuracy of the knowledge representations and multi-task utilization.}
    \label{tbl:preft_llm}
    \begin{tabular}{cccc}
    \hline
         \textbf{Model} & \textbf{Anchor Model} & \textbf{Data Source} & \textbf{Task} \\
    \hline
         ChiMed-GPT \cite{ChiMed-GPT} & Ziya-13B-v2 \cite{Ziya-13B-v2} & CMD. & IE, QA, MD \\
         Qilin-Med & Baichuan-7B & ChiMed & MT, QA \\
    \hline
    \end{tabular}
\end{table}

\textbf{ChiMed-GPT:} Tian et al. proposed a new benchmark LLM designed for the Chinese medical field, based on Ziya-13B-v2, inheriting its ability to handle a wide range of context lengths and extending the context length to 4096 tokens. ChiMed-GPT integrates pre-training, supervised fine-tuning (SFT), and reinforcement learning from human feedback (RLHF) stages, ensuring that it not only captures domain-specific knowledge but also adapts to various situations, surpassing existing models that typically rely solely on SFT. Evaluations on practical tasks (including IE, QA, and dialogue generation) show that ChiMed-GPT performs superiorly on general-domain LLMs. However, researchers have found that the model exhibits potential biases that urgently need to be addressed.

\textbf{Qilin-Med:} Ye et al. have constructed a subsample dataset ChiMed, which includes medical question answering, plain text, KGs, and dialogue, among others. Based on the characteristics of the data, they proposed a multi-stage training method that combines domain-specific continued pre-training (DCPT), SFT, and direct preference optimization (DPO). The final experiments demonstrate that, during the CPT and SFT stages, the Qilin-Med trained using the multi-stage training method achieved accuracy rates of 38.4\% and 40.0\% on the CMExam, respectively, surpassing the baseline model Baichuan-7B’s 33.5\%. In the DPO stage, on the Huatuo-26M test set, Qilin-Med scored 16.66 and 27.44 on the BLEU-1 and ROUGE-1 metrics, respectively, better than SFT’s 12.69 and 24.21. Although the training method of Qilin-Med shows clear advantages in improving the performance of LLMs for medical applications, it still has certain limitations in terms of datasets, multi-stage training preferences, and evaluation metrics. For instance, although the dataset covers comprehensive medical knowledge, it mainly focuses on Chinese medical knowledge, limiting the model’s global applicability. The preference risk in the multi-stage training process may introduce the preferences of human evaluators, and metrics like BLEU and ROUGE may not be able to capture the entire performance of the model in medical scenarios.

\subsubsection{Building Medical LLMs from Multimodal Data}
Compared to aforementioned monomodal medical LLMs, multimodal medical LLMs contain more comprehensive knowledge that contributes to more accurate predictions. Moreover, multimodal data contains knowledge with smaller granularity, which can compensate for the deficiencies of a single modality. Therefore, multimodal LLMs can help extract common features across modalities. 

In the following section, we will conduct an in-depth analysis of the current efforts in the field, presenting the current research status and development trends using the construction methods of strategies such as pre-training and fine-tuning. Table \ref{tbl:mutil_llm} summarises the construction methods of multimodal medical LLMs.

\begin{table}[h]
	\caption{Summary of methods for constructing multimodal medical LLMs. We present these methods according to the modality of their training strategies.}
    \label{tbl:mutil_llm}
	\centering
 \resizebox{\textwidth}{33mm}{
	\begin{tabular}{ccccc}
        \hline
		\textbf{Strategy} & \textbf{Model} & \textbf{Anchor Model} & \textbf{Data Source} & \textbf{Task} \\
	\hline
        Pre-training & BioMedGPT \cite{BioMedGPT} & LLaMA2-Chat-7B & S2ORC \cite{S2ORC} & \makecell{BioMedical QA\\Molecule QA\\Protein QA} \\
        \hline
        \multirow{13}{*}{Fine-tuning} & Med-PaLM M & PaLM-E \cite{PaLM-E} & MultiMedBench & \makecell{QA\\RS\\VQA\\RG\\MIC} \\
        \cline{2-5}
             & Qilin-Med-VL & \makecell{Clip-ViT-large-patch14-336\\ChineseLLaMA2-13B-Chat} & ChiMed-VL & VQA \\
        \cline{2-5}
             & LLaVA-Med \cite{LLaVA-Med} & GPT-4 \cite{GPT-4} & PMC-15M & VQA \\
             \cline{2-5}
             & XrayPULSE & PULSE & \makecell{MIMIC-CXR\\OpenI} & Multimodal MD\\
             \cline{2-5}
             & XrayGLM \cite{XrayGLM} & VisualGLM-6B & \makecell{MIMIC-CXR\\OpenI} & \makecell{Medical Imaging diagnosis\\MD} \\
             \cline{2-5}
             & Visual Med-Alpaca \cite{VariousMedQA} & LLaMA-7B & VariousMedQA \cite{VariousMedQA} & \makecell{QA\\VQA\\MD} \\
        \hline
\end{tabular}
}
\end{table}

\begin{figure}[htp]
    \centering
    \includegraphics[width=.8\textwidth]{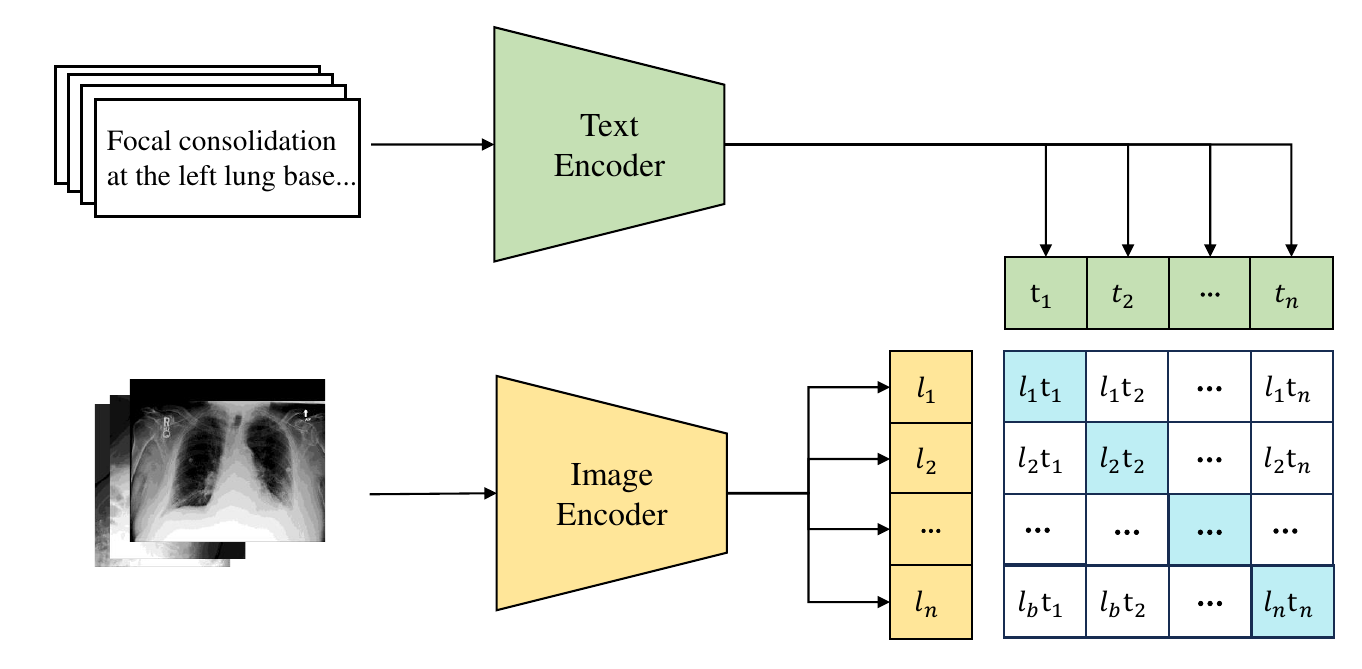}
    \caption{Contrastive learning multimodal pre-training technique. This type of pre-training has strong transferability and zero-shot capabilities. }
    \label{clip}
\end{figure}

\paragraph{Pre-training} Building multimodal LLMs often requires creating uniform multimodal representations of the same object across different modalities through pre-training methods. Traditional multimodal pre-training methods usually employ contrastive learning (CL) to help models distinguish between different modal representations. As shown in Figure \ref{clip}. The core idea is to teach a model's representation by comparing the similarity between positive and negative samples. This is done even if the distance between positive examples becomes closer and the distance between negative examples becomes farther. Models \cite{MedCLIP} created with CL generally perform better for multimodal image-text retrieval.

\begin{figure}[htp]
    \centering
    \includegraphics[width=.8\textwidth]{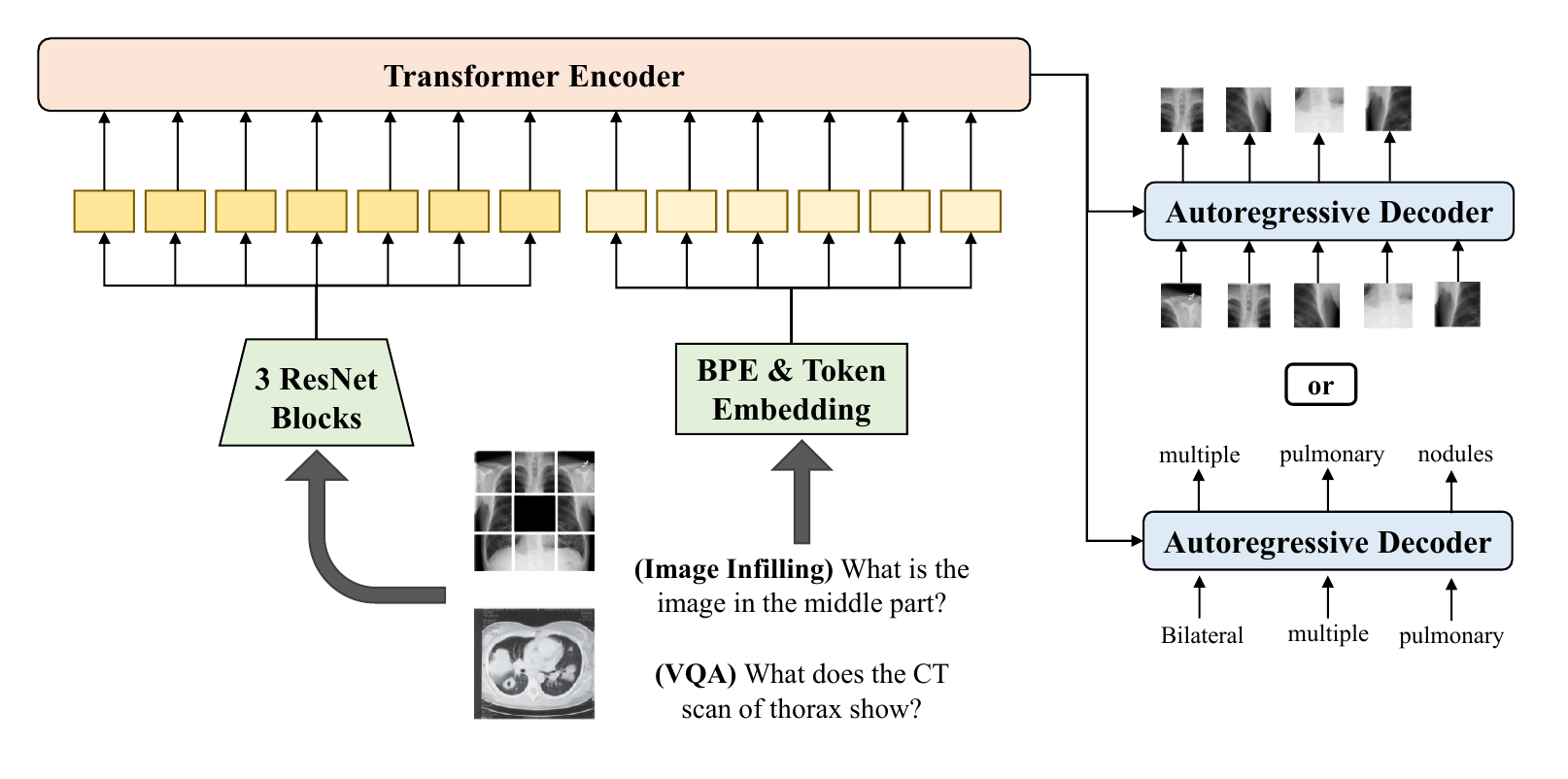}
    \caption{An end-to-end multimodal pre-training technique. This type of pre-training views the model as a pipeline solution for specific tasks. Therefore, their abilities can be transferred among similar tasks easily.}
    \label{figure2}
\end{figure}

Once the LLM is proposed, pre-training the model using full parametric CL methods becomes challenging. This is because pre-training typically necessitates a substantial amount of high-quality data, and full parametric training is hardware-intensive and costly. Therefore, some researchers have proposed new training approaches, as shown in Figure \ref{figure2}.

For example, Luo et al. \cite{BioMedGPT} proposed a novel end-to-end multimodal semantic understanding framework called BioMedGPT. This framework uses the pre-trained large language model BioMedGPT-LM, specifically designed for the biomedical domain, to connect natural language, biological encoding language, and chemical molecular language. The text, molecules, and proteins are encoded separately using encoders in the model. The feature vectors of the pairs are then concatenated into the input for the model's encoder. Finally, a self-regressive decoder is employed to generate results for each modality.
Consequently, the model can support interactive QA across natural language and molecular language modalities. It has potential applications in drug target exploration and mining, lead compound design and optimization, and protein design. Additionally, it has achieved SOTA performance on multiple biomedical QA benchmark datasets, including USMLE, MedMCQA, and PubMedQA, matching the level of human medical experts. It has also successfully passed the United States Medical Licensing Examination.

\paragraph{Fine-tuning} Similar to constructing large medical text models, researchers prefer to fine-tune model parameters to improve the model's performance on specific tasks, which incurs a smaller training cost. This approach to fine-tuning can be divided into three main categories: end-to-end methods, BLIP-derived methods, and prompt-combination methods.

\textbf{End-to-end Methods.}
The end-to-end approach is comparable to the pre-training approach of BioMedGPT mentioned earlier. This involves encoding data from different modalities separately and then aligning the multimodal vector representation using a unified model. Med-PaLM M is a typical example of this approach.

Med-PaLM M is a multimodal generative model developed by Google Research and DeepMind for encoding and interpreting biomedical data, including clinical language, medical imaging, and genomics, among others. The model is fine-tuned to align the PaLM-E to the biomedical domain through the MultiMedBench benchmark. Med-PaLM M achieves competitive or superior performance to the SOTA in all tasks of MultiMedBench. Additionally, Med-PaLM M significantly outperforms PaLM-E, demonstrating the importance of biomedical fine-tuning and alignment. The experiments on clinical adaptability also show the potential clinical utility of the model. Additionally, it demonstrates zero-shot generalization to medical concepts and tasks, indicating the model's ability to reason and make decisions for medical situations for which it has not been explicitly trained.

\begin{figure}[htp]
    \centering
    \includegraphics[width=.8\textwidth]{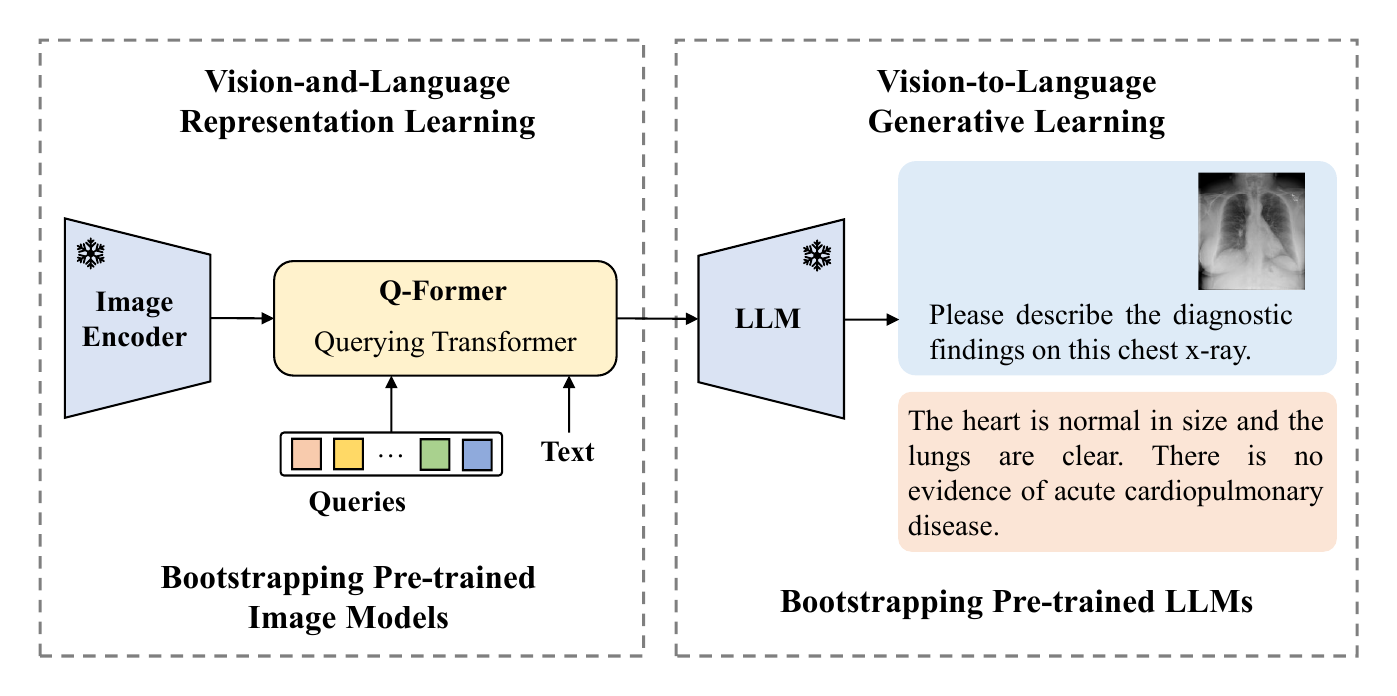}
    \caption{BILP fine-tuning method. This type of method treats different parts of the model separately so that we can improve the model performance without impacting the model generalization.}
    \label{figure3}
\end{figure}

\textbf{BLIP-derived Methods.}
However, the end-to-end approach lacks the flexibility to solely utilize pre-trained models that support multimodality or combinations of already trained multimodal encoders. To address this issue, the researchers have utilized the BLIP training framework, which offers greater flexibility in combining LLMs that support either image processing or text analysis. As shown in Figure \ref{figure3}, the approach integrates a Q-Former neural network between the image encoder and text model. The multimodal models can be linked by training the Q-Former alone while keeping the image and text models frozen.

The framework introduces three main learning objectives for training tasks: image-text matching, image-based text generation, and image-text CL. In the medical field, the models constructed using this method are mainly:

\begin{itemize}
    \item \textbf{Qilin-Med-VL}: Liu et al. proposed the Qilin-Med-VL, the first Chinese large-scale visual language model that integrates textual and visual data analysis. The model combines a pre-trained visual transformer (Clip-ViT-large-patch14-336) with a base LLM (ChineseLLaMA2-13B-Chat) to improve the ability to generate medical text and answer complex medical queries. Additionally, Liu et al. released the ChiMed-VL dataset, which consists of over 1 million image-text pairs. This dataset has been meticulously crafted to offer a thorough and complete analysis of medical information through diverse image types.
    \item \textbf{LLaVA-Med}: To address the lack of complexity in the understanding and dialoguing of biomedical images by general-domain vision-language models, Li et al. have proposed a cost-effective method for training a vision-language dialogue assistant, LLaVA-Med, capable of answering open research questions in biomedical images. This is the first attempt to extend multimodal instruction tuning to the biomedical field, enabling end-to-end training for developing a biomedical multimodal dialogue assistant. The model is based on LLaMA, fine-tuned on image-text pairs to align with biomedical vocabulary, and then further trained with instruction data automatically generated by GPT-4 based on sampled biomedical image-text pairs from PMC-15M, to learn open-ended dialogue semantics. Experimental studies finally validate the effectiveness of domain-specific instruction tuning. On established biomedical VQA datasets, the performance of fine-tuned LLaVA-Med generally surpasses that of supervised SOTA methods.
    \item \textbf{XrayPULSE}\footnote{https://github.com/openmedlab/XrayPULSE}: XrayPULSE is an extension of PULSE, a model that uses MedCLIP as a visual encoder and a simple linearly transformed Q-former (BLIP2) as an adapter to incorporate the image into PULSE. To align frozen visual encoders and LLMs via adapters, OpenMEDLab generates the Chinese version of Xray-Report paired data from radiological reports (MIMIC-CXR and OpenI) to create the Chinese version of Xray-Report paired data. Finally, XrayPULSE has been fine-tuned on this dataset to serve as a biomedical multimodal dialogue assistant, extending PULSE.
    \item \textbf{XrayGLM}: The XrayGLM proposed by Wang et al. is the first Chinese multimodal medical LLM for chest X-rays, demonstrating extraordinary potential in medical image diagnosis and multi-turn interactive dialogue. The base model, VisualGLM-6B, is trained via the BILP method based on ViT and ChatGLM2. In the fine-tuning stage, the model was trained on the publicly available chest X-ray dataset MIMIC-CXR and a medical multimodal dataset constructed with X-ray images and diagnostic reports, aided by ChatGPT and publicly available datasets.
\end{itemize}

\textbf{Prompt-combination Methods.}
In addition to the previously mentioned multimodal construction mechanism, another way to construct multimodal models is to link already trained multimodal encoders through prompts. Figure \ref{figure4} below. This method is cheaper and more convenient for training than the abovementioned methods.

\begin{figure}[htp]
    \centering
    \includegraphics[width=.7\textwidth]{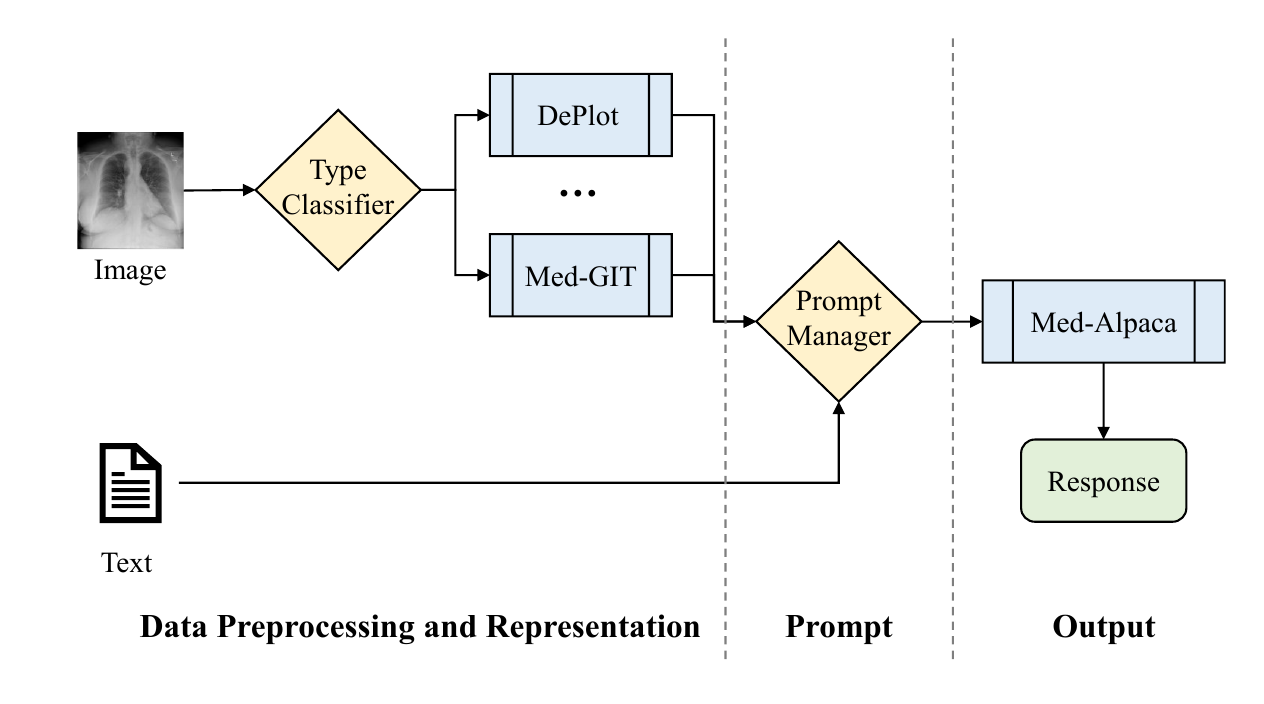}
    \caption{Prompt-combination fine-tuning method. This type of method combines the excellent performance of models over multiple modalities to create a stronger model through prompt design.}
    \label{figure4}
\end{figure}

Visual Med-Alpaca is a prime example of a multimodal healthcare LLM constructed with prompts. The model is based on LLaMa-7B and is trained using an instruction set developed collaboratively by GPT-3.5-Turbo and human experts. The instruction set extracts medical questions from medical datasets in the BigBIO repository. These questions are then used to direct GPT-3.5-Turbo to generate answers. The resulting QA pairs undergo multiple human filtering and editing rounds to optimize their quality. The final instruction set contains 54k high-quality instructions. Finally, with hours of fine-tuning and a plug-and-play vision module, Visual Med-Alpaca can perform various medical tasks.

In summary, these two building mechanisms of LLMs above have shown excellent performance in one or more medical tasks. However, they still face several challenges, including quality issues in data annotation, uneven data distribution, insufficient model interpretability and transparency, and limitations in multimodal information fusion. During the evaluation of model performance, most methods rely primarily on automated medical task evaluation metrics, while relatively few incorporate manual evaluation. Furthermore, the practical application value of these models still requires further validation and evaluation.

Although these LLMs are not yet fully ready for practical application, their great potential cannot be ignored. They offer useful references and insights for practical applications in the medical field and indicate directions for future research and development.

\subsubsection{Biomedical Agent}
\begin{figure}[htp]
    \centering
    \includegraphics[width=.8\textwidth]{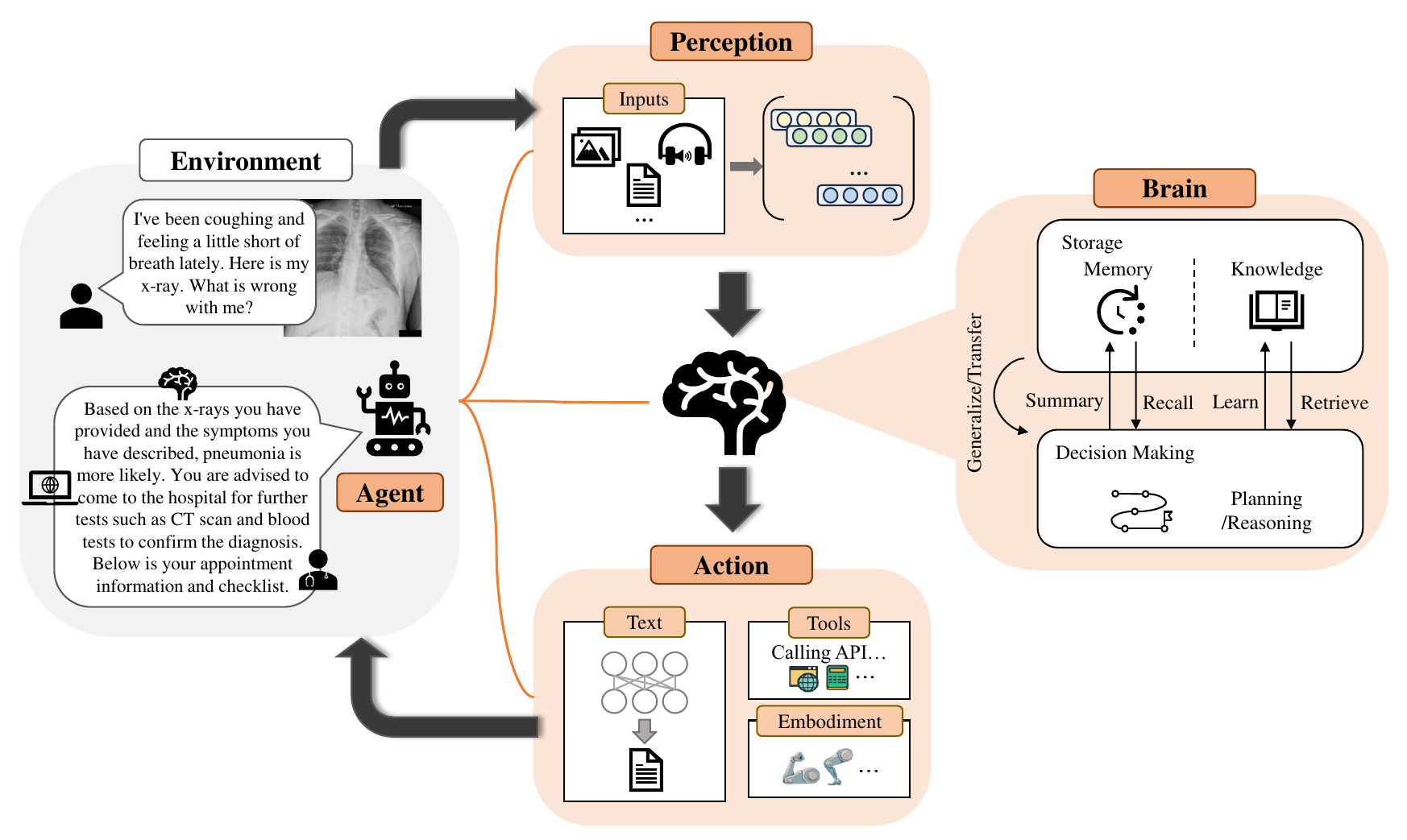}
    \caption{LLM-based agent framework. It contains three components: brain, perception, and action.}
    \label{agent}
\end{figure}

In addition to the traditional approach of pre-training and fine-tuning to develop medical LLMs, the ability of complex reasoning and tool invocation possessed by LLMs has led to an increasing interest in using agents, which are LLM derivatives. Agents aim to design and build computer-based agents that exhibit intelligent behavior. There is currently no widely accepted definition of an agent. However, it is generally understood \cite{Xi23TheRise} that artificial agents are artificial entities that can perceive their surroundings using sensors, make decisions, and take responsive actions using actuators, as shown in Figure \ref{agent}.

Integrating LLMs through agents enables the automation of complex tasks using LLMs. These agents can interact with the environment, humans, or other agents through perception, reasoning, and planning \cite{Xi23TheRise} \cite{Zhou23Agents} \cite{Song23LLM-Planner}. Recent work \cite{Cho23AnIntegrative} has shown the potential of combining agents with LLMs, advancing healthcare automation.

To facilitate online communication with adolescents with autism spectrum disorders, Ali et al. \cite{Ali20AVirtual} proposed an online virtual dialogue agent. Users engage in casual conversations with the virtual agent, receiving real-time feedback and summaries. The agent generates feedback and dialogue automatically using hidden Markov models and a pattern-driven dialogue manager capable of handling multi-topic conversations. This system is expected to help adolescents alleviate autistic symptoms and positively impact behavior change.

Abbasian presented a conversational health agent (CHAs) framework \cite{Abbasian23Conversational} based on an LLM that enables CHAs to generate personalized responses to users' health queries. This framework enhances personalized medical healthcare services by participating in emotionally charged dialogues and processing multimodal data in an interactive system. Integrating healthcare data sources, supporting multilingual and multimodal dialogue, and interacting with various user data analysis tools provide critical thinking, knowledge acquisition, and problem-solving capabilities. The framework aims to improve the ability of CHAs to respond effectively to healthcare queries by analyzing input questions, collecting data, performing operations, and providing personalized responses.

To address LLMs' challenges in handling domain-specific terminology and reasoning with expert knowledge, Tang et al. \cite{Tang23MedAgents} proposed a multidisciplinary collaboration framework. In the medical domain, this framework uses role-playing agents based on LLMs to engage in iterative multi-turn discussions, thereby enhancing LLMs' knowledge and reasoning capabilities. The framework comprises five key steps: gathering domain expertise, proposing individual analyses, summarising these analyses in reports, iterating in discussions until consensus is reached, and finally, making decisions. Experimental results on nine datasets (including MedQA, MedMCQA, PubMedQA, and six subtasks from MMLU) demonstrate the exceptional performance of the framework in mining and exploiting medical expertise within large models and extending their reasoning capabilities.

Using agents allows LLMs to more accurately use external tools for medical and biological domain expertise, thereby mitigating the problem of hallucination that LLMs face in certain domains. Jin et al. \cite{Jin23GeneGPT} presented GeneGPT, a method that guides LLMs to use the National Center for Biotechnology Information (NCBI) Web API to answer genomics questions. Specifically, through context learning and an improved decoding algorithm, this method encourages large models to use external tools to solve related judgments by detecting and executing API calls. Experimental results show that GeneGPT achieves SOTA performance compared to large models such as Bing, BioMedLM, BioGPT, GPT-3, and ChatGPT on eight tasks of the GeneTuring benchmark.

To improve the effectiveness of LLMs in clinical tasks, Alex et al. \cite{Alex23OpenMedCalc} investigated the ability of ChatGPT to perform medical calculations and evaluated its performance on 48 different clinical calculation tasks. The study results showed that ChatGPT performed inconsistently in clinical calculations, providing inaccurate answers in a third of these trials. To address this issue, the study developed an open-source clinical computation API. It integrated it into ChatGPT, allowing the LLM to interact with external programs and tools the same way as an agent. The results show that the augmented model significantly improves accuracy by evaluating performance on three common clinical computing tasks using 75 clinical cases and comparing it to other models.

These models showcase the potential of applying AI to agents in various domains, from aiding social interactions for individuals with autism spectrum disorders to offering personalized healthcare guidance. However, some unresolved issues remain, such as improving the handling of complex contexts and enhancing the naturalness of dialogues in virtual conversational agents. Additionally, there is a need to more efficiently integrate multimodal data and improve the accuracy of personalized responses in the framework of health agents. These areas require further investigation.

\subsection{Collaborative Development of Medical LLMs and Clinical KGs}
In recent years, with the rapid development of KGs and LLMs, combining KGs and LLMs has become a popular research direction \cite{LLM+KG}.

LLMs can adapt to different contexts and generate natural language for various NLP tasks, including text summarisation, QA systems, and machine translation. However, it suffers from problems such as factual fabrication and lack of interpretability due to being parameterized implicit knowledge. In contrast, KG is structured explicit knowledge that is naturally interpretable. Domain-specific knowledge is of high quality but has high construction overheads, is often incomplete, and lacks NLP capabilities. Therefore, combining the two can effectively compensate for their deficiencies, take advantage of their complementary strengths, and jointly improve the NLP capability and application scope.

In medical AI research, we categorize the integration of LLM and KG into three types: (1) \textbf{LLMs for Medical KGs}: LLM empowers medical KG construction, which means using LLM’s advantage to enrich and optimize medical KGs. (2)\textbf{ KGs for Medical LLMs}: Medical KG empowers LLM, integrating structured knowledge from KGs into LLM and enhancing its knowledge understanding and application capabilities in specific fields. (3) \textbf{Medical LLMs and KGs}: The mutual synergy between medical KG and LLM is to achieve more efficient knowledge retrieval and language generation through their interaction. Next, we will introduce the relevant research progress in detail.

\subsubsection{LLMs for Medical KGs}
Constructing and maintaining traditional KGs is time-consuming and requires much manual input. Each step, from data collection and cleaning to knowledge collation and annotation, as well as regular updating and iteration, requires many annotators and professional knowledge support. This process is costly and inefficient. LLM provides an effective solution for automating knowledge acquisition on KG platforms. It has excellent generalization capabilities and can process textual information. LLMs are used for constructing medical KGs in three main types: method paradigm extension, processing object extension, and task type extension.

\paragraph{Method Paradigm Extension} In healthcare, method paradigm extension refers to combining LLM with KGs to extract, represent, and reason about medical data and knowledge using techniques such as graph neural networks (GNNs) and NLP. This approach can help medical LLMs better understand and apply medical knowledge. Research in this direction has:

\begin{itemize}
    \item \textbf{ERNIE-Health}\footnote{https://huggingface.co/nghuyong/ernie-health-zh}: ERNIE-Health, leveraging Baidu Wenxin's knowledge-enhanced pre-trained language model ERNIE, acquires a vast amount of medical knowledge through technologies that enhance medical knowledge. ERNIE-Health has learned over 600,000 medical terms and over 40 million medical QA data, significantly outperforming other compared models on the CBLUE multi-tasking list (which covers five major tasks: medical information extraction, medical term normalization, medical text classification, medical sentence relation judgment, and QA). This greatly enhances the model's understanding and modeling capabilities of medical professional knowledge. As a result, the model can accurately identify medical terms within a vast amount of medical text, thus automating the extraction of medical knowledge.
    \item \textbf{GatorTron} \cite{EHRs} \cite{GatorTron}: GatorTron is an early LLM developed for healthcare specifically to explore how clinical LLMs with tens of billions of parameters can derive knowledge from unstructured EHRs. The model was trained on over 9 billion data tokens, including more than 8.2 billion words of non-identifying clinical text. It was systematically evaluated on five clinical NLP tasks: clinical concept extraction, medical relationship extraction, semantic text similarity, medical natural language inference, and MQA. The final experimental results demonstrate that GatorTron significantly improves sentence-level and document-level NLP tasks. Thus, the model accurately represents medical knowledge, making it an important tool for medical KG vectorization.
    \item \textbf{Graph-ToolFormer} \cite{Graph-ToolFormer}: Zhang et al. proposed Graph-ToolFormer, a framework focused on graph inference that teaches LLMs to use external graph inference tools through ChatGPT-enhanced prompts. During training, the model instructs Graph-ToolFormer to handle various graph data reasoning tasks, including basic graph data loading and graph attribute reasoning in the field of biochemical molecules. When compared to other LLMs on reasoning tasks, it is evident that Graph-ToolFormer improves LLM's capacity to apply knowledge and reasoning-related aspects.
    \item \textbf{KSL} \cite{KSL}: To use the reasoning capabilities of LLMs in assisting with retrieving knowledge from external KBs, researchers have also explored the use of additional modules, such as GNNs. Feng et al. proposed the Knowledge Solver (KSL), which teaches additional networks to search for basic knowledge from external KBs by exploiting the strong generality of LLMs themselves. Specifically, the model transforms retrieval into a multi-hop decision sequence, enabling LLMs to search for knowledge in a zero-shot task. Additionally, the KSL provides complete search paths, increasing the interpretability of the LLM reasoning process. Experiments on three datasets, including MedQA-USMLE in healthcare, demonstrate that the approach significantly improves the benchmark performance of the LLMs.
\end{itemize}

\paragraph{Processing Object Extension} In healthcare, processing object extension refers to expanding the application of LLM to additional types of medical data and knowledge, such as medical records, medical images, and drug information. This approach can enhance the effectiveness of medical LLM applications.

For instance, BioMedGPT is a fundamental model in the multimodal biomedical field that aims to improve the capability of various downstream tasks by unifying the representation learning of molecules, texts, and knowledge. BioMedGPT integrates diverse and heterogeneous data at the data level, including genes, molecules, cells, proteins, literature, patents, and KBs. For the first time, it introduces knowledge into the model construction, enabling unified representation learning of biological text and knowledge, which enhances the model’s generalization ability and interpretability. Regarding application tasks, BioMedGPT can handle multiple tasks such as natural language, drug property prediction, and cross-modal generation, allowing for a comprehensive exploration of all tasks in the field of life sciences. It has achieved the best results in several key downstream tasks.

\paragraph{Task Type Extension} In healthcare, task type extension refers to expanding the application scenarios of LLM to include more medical tasks, such as disease diagnosis, treatment recommendations, and medical research. This approach can enhance the flexibility and applicability of medical LLM. For instance, ChatDoctor is built on the LLaMA model, which has been trained and fine-tuned using over 11 million real doctor-patient dialogues. It is also equipped with an external KB containing information on more than 700 diseases. By leveraging the excellent generative and multi-tasking generalization capabilities of the LLM, ChatDoctor can provide patients with accurate responses to their queries, along with personalized treatment recommendations from reliable sources.

\subsubsection{KGs for Medical LLMs}
While ChatGPT, GPT-4, and other LLMs have shown impressive language comprehension and generation abilities, they are still limited by their pre-trained corpus and model capabilities. They also face challenges such as limited knowledge memory, weak knowledge reasoning, and difficulty in updating knowledge. To address these issues, KGs can be a viable solution due to their structured and explicit representation of knowledge, natural interpretability, and high-quality information. Additionally, KGs can also represent and generate Chain-of-Thought (CoT), which can further enhance the reasoning ability of LLMs by structuring better CoT. The approaches with KGs for medical LLMs can be broadly classified into three categories: knowledge learning during the training, knowledge integration during the reasoning, and continuous learning from historical experience.

\paragraph{Knowledge Learning During the Training} In healthcare, knowledge learning during the training mainly involves training the LLM with a large amount of data from medical texts and KGs to enable it to understand professional knowledge and concepts. The key to this phase is to ensure the training data's quality and diversity, the training task's rational design, and the loss function to improve the LLM's ability to understand and express medical knowledge.

\begin{itemize}
    \item \textbf{BenTsao} used the LLaMA-7B model along with Chinese medical instruction-tuning technology to create an efficient medical LLM. Using the medical KG and GPT3.5 API, they created a dataset of Chinese medical instructions. They then fine-tuned LLaMA's instructions based on this dataset, significantly improving the model's QA performance. The model is trained using CMeKG and other relevant Chinese medical literature resources. After a series of designs and optimization, a large-scale medical LLM named "BenTsao" has been created. The model outperforms comparative models like LLaMA and Alpaca in the MQA task.
    \item \textbf{ShenNong-TCM-LLM}\footnote{https://github.com/michael-wzhu/ShenNong-TCM-LLM} generates data using the open-source TCM KG and is trained using the entity-centric self-instruct method. It outperforms other instruction-tuning models on the cMedKnowQA dataset.
    \item \textbf{DISC-MedLLM} \cite{DISC-MedLLM} is a medical LLM designed specifically for healthcare conversational scenarios. The model combines medical KGs, real-world dialogues, and human feedback methods to create a high-quality supervised fine-tuning dataset. This dataset is then used to train the model on the Baichuan-13B-Base. Finally, experiments on a large medical dataset validate the significant impact of the model in improving the medical consultation capabilities of LLM.
\end{itemize}

\paragraph{Knowledge Integration During the Reasoning} In healthcare, knowledge integration during reasoning focuses on combining LLMs with medical KGs to answer medical questions or provide medical advice through reasoning and querying. The key to this phase is to design effective reasoning algorithms and query strategies to fully utilize the information in the medical KG and improve the LLM's reasoning and answering capabilities.

Gao et al. \cite{Gao23Leveraging} proposed an innovative approach to improve LLMs for automatic diagnosis generation by introducing medical KGs and graph-based models. This approach eliminates the need for pre-training and instead uses KGs to aid in explaining and summarising complex medical concepts during inference. Experimental results on a real hospital dataset show that the method effectively improves the accuracy of automatic diagnosis generation.

There has been less research in this direction in healthcare, but Sun et al. \cite{LLM-KG} have proposed an LLM-KG integration paradigm that could potentially be extended to medical LLMs. This paradigm considers the LLM as an agent that interactively explores relevant entities and relationships on the KG and performs reasoning based on the retrieved knowledge. The paradigm is achieved by introducing a new approach called Think-on-Graph (ToG). The LLM agent performs an iterative bundle search on the KG, discovers the most promising inference paths, and returns the most probable inference results. This approach achieves SOTA performance on several KBQA datasets.

\paragraph{Continuous Learning from Historical Experience} In healthcare, continuous learning from historical experience focuses on continuously optimizing and improving the performance of LLM by recording and analyzing historical data and cases. The key to this phase is to design effective methods for collecting and analyzing experiences and cases to help LLMs continuously learn and improve. Among the above models, models such as DISC-MedLLM and HuaTuoGPT perform well in multi-turn dialogue scenarios, partly due to the continuous learning and analysis of historical dialogue data.

\subsubsection{Medical LLMs and KGs}
In the field of AI applications within healthcare, the demand for interpretability, credibility, and traceability of outcomes exceeds that of other domains. This necessitates a combination of medical KGs with various LLM technologies to play a more significant role. LLMs, often due to their black-box nature, may produce biased results. However, KGs contribute to interpretability, credibility, and traceability, aiding in understanding the workings of LLMs. Therefore, integrating KG throughout the entire lifecycle of LLMs, from pre-training to various application stages, is an efficient strategy. This integration not only enhances the training efficacy of LLMs but also improves the practicality and reliability of their inferential results. Nevertheless, development in this area is still somewhat lacking.

Currently, medical KGs and LLMs are constructed collaboratively using various methods such as information filtering, retrieval augmented generation by KG, generation enhanced extraction, and dynamic collaborative enhancement.
\begin{itemize}
    \item \textbf{Information Filtering.} Information filtering refers to extracting relevant knowledge from KBs and then sending it to the LLM for answering. Zhang et al. \cite{LangChain} proposed using LangChain to create a new model that combines KGs and LLMs deeply. Firstly, the input text of questions related to TCM formulas undergoes information filtering, specifically text classification, to determine its relevance. Secondly, LangChain retrieves knowledge related to the text from the KB and inputs it into LLMs such as ChatGPT and ChatGLM, along with the question in the form of prompts. The LLM then generates a professional answer through reasoning. Finally, the answer undergoes knowledge extraction to extract the triples. The extracted triple is then matched with the existing prescription KG to verify its expertise. Furthermore, the nodes in the KG are used as input for the LLM to obtain natural language explanations, achieving bidirectional conversion between the LLM and the KG.
    \item \textbf{Retrieval Augmented Generation by KG.} To generate more knowledgeable responses based on LLM and a self-built KB, collaboration between LLM and KG can also be achieved. Soman et al. \cite{KG-RAG} introduced a task-agnostic KG-based retrieval-augmented generation (KG-RAG) framework, which leverages large biomedical KGs like SPOKE in conjunction with LLMs (e.g., Llama2, GPT-3.5, and GPT-4) to generate meaningful biomedical texts based on established knowledge. This framework combines explicit and implicit knowledge from KG and LLM, respectively, thereby enhancing the adaptability of general LLMs to address specific domain problems within a unified framework. KG-RAG continuously improves the performance of LLMs on different task types, including drug-use queries, biomedical questions, and multiple-choice questions.
    \item \textbf{Generation Enhanced Extraction.} By reinforcing the information extraction capability of LLM, the generated natural language responses are extracted to structured knowledge and matched with the professional KG for verification, realizing a deep integration method between LLMs and KGs. Xu et al. \cite{ClinGen} conducted in-depth research on using LLMs for clinical text generation tasks and proposed an innovative and resource-efficient method called ClinGen, which integrates knowledge into the entire clinical process. This model utilizes a medical domain-specific KG and an LLM to guide data generation. Research on seven clinical NLP tasks and 16 datasets shows that ClinGen can continuously improve performance and significantly enrich the diversity of generated instances in various tasks.
    \item \textbf{Dynamic Collaborative Enhancement.} Unlike previous approaches that synergize the LLM with static KGs, Li et al. \cite{dalk} propose a DALK framework that combines the LLM with dynamic KGs to improve the model's ability to answer Alzheimer's disease (AD) questions. The approach first utilizes the LLM to construct an evolving AD-specific KG from AD-related scientific literature, then enhances the inference of the LLM by selecting appropriate knowledge from the KG using a coarse-to-fine sampling method with a novel self-aware knowledge retrieval approach, and finally achieves mutual enhancement between the LLM and the KG. DALK also performs well in a specially constructed ADQA benchmark, demonstrating its effectiveness within its domain of expertise.
\end{itemize}

\section{Applications in Medicine}
\label{chp: 4}
In addition to the scientific research progress summarized above, medical LLMs that integrate clinical knowledge have been applied in various scenarios, including medical research, drug research and development (R\&D), intelligent diagnosis and treatment (D\&T), medical equipment operation and maintenance, and hospital management. 

These applications can assist medical development in terms of diagnosis, medical imaging, drug development and innovation, medical dialogue services, and personalized treatment plans, which require knowledge of Chinese medicine, pharmaceutical molecules, biomedicine, and genomics in multilingual contexts. To better understand medical LLM applications, we classify them into two types based on different modalities: plain-text and multimodal.

\subsection{Industrial Text Medical LLMs}
Table \ref{tbl:in_cor_app} presents successful cases of transitioning text-based medical LLMs from academic to industry:

\textbf{UniGPT-Med}: UniGPT-Med, based on UniGPT, is specifically designed to understand medical data and knowledge. It is trained on up to two trillion tokens, including a knowledge graph containing 1.6 million concepts, 3.7 million terms, and 8.4 million relationships. Unisound has successfully developed a medical record generation and quality control system using this technology. These related products have been implemented in hospitals to help medical professionals diagnose diseases using AI, improve the efficiency of outpatient medical record entry, and save time while enhancing the accuracy of diagnosis and treatment.

\textbf{EyeGPT}: EyeGPT, developed by Wenzhou International Optometry Innovation Centre for ophthalmology, is pre-trained based on many natural and medical professional corpus and fine-tuned for medical scenarios using professional data such as ophthalmology EHR information. Its application is mainly used in scientific research and clinical medical assistance by integrating NLP, computer vision (CV), and other technologies to provide intelligent assistance tools for medical professionals and patients with interactive questions and answers.

\textbf{DaJing TCM}: DaJing TCM has released the QiHuangWenDao Model, which focuses on TCM and aims to achieve intelligent clinical diagnosis, treatment, and health regulation. Based on the KG, diagnosis, and treatment knowledge, the model is trained using a four-layer progressive training approach of pre-training, supervised fine-tuning, rewarding model, and reinforcement learning. Currently, 1,704,605 TCM experts have participated in project evaluation. Additionally, DaJing TCM has released two sub-models: the Medical Model for clinical diagnosis and treatment and the Health Model for recommending TCM health and conditioning programs.

\textbf{DingDangKuaiYao}: Dingdang Health has launched applied pharmaceutical AI products - Dingdang Pharmacist and Dietitian AI Assistant - based on its self-developed HealthGPT. These AI products are integrated into the DingdangKuaiYao app. The company's offline physical chain smart pharmacies operate in 19 cities across the country, utilizing an integrated online and offline operations model. Through its three core businesses of Fast Medicine service, online health consultation, and chronic disease and health management, DingdangKuaiYao provides users with convenient and beneficial health services such as 7×24 instant health support and online consultations with professional doctors and pharmacists.

\textbf{Gushengtang}: Gushengtang, in collaboration with Baidu Lingyi, has jointly developed the Gushengtang TCM model to provide intelligent TCM services like diagnosis and treatment, and formula recommendations. This model integrates a wealth of clinical EHRs, case information, and KGs from renowned Chinese medical masters. The products derived from this model are divided into the patient and doctor ends. The patient end provides various services such as disease consultation, quick finding of doctors, intelligent department guidance, medication guidance, and intelligent customer service through a dialogue mode, while the doctor end mainly provides digital empowerment for clinical diagnosis and academic research for doctors. 

\textbf{Xunfei Xiaoyi}: The Xinghuo Medical Model is based on the Xinghuo Cognitive Model V3.0, aiming to create an AI health assistant for everyone. It has billions of high-quality medical data, allowing the model to master professional medical knowledge further. Based on this, the Xunfei Xiaoyi APP and mini program have been released, providing services such as medical health knowledge inquiry and Q\&A, medical strategy, medication inquiry, report interpretation, TCM syndrome differentiation, and health record management. These services aim to assist users in improving medical efficiency and experience, allowing the inclusive light of AI healthcare to shine into every household and become everyone's AI health assistant.

\textbf{Zhiyun Health}: Zhiyun Health, based on ClouD GPT, is mainly used in hospitals, Internet hospitals, and other scenarios. It has been implemented in the medical application scenarios of intelligent cloud AI-assisted diagnosis and AI drug and device research and development. In hospital product applications, ClouD GPT provides AI-assisted clinical diagnosis and treatment modules. In pharmacy product applications, ClouD GPT supports doctors and pharmacists in online diagnosis and treatment service management. Currently, Zhiyun Health has made rapid progress in the research and development of AI-assisted medical devices. Product development for diabetes, coronary heart disease, hyperlipidemia, and other areas will gradually complete clinical validation, providing core technical solutions for digital chronic disease management.

\textbf{WiNEX}: WiNGPT is a medical vertical-oriented LLM developed by Wining Health. Based on the general LLM, WiNGPT strengthens the model capability by obtaining high-quality training data through data engineering based on public and medical domain data. Based on WiNGPT, WiNEX Copilot Healthcare Intelligent Assistant has been developed to provide intelligent services for healthcare professionals in the areas of medical record document generation, medical quality supervision, risk monitoring and early warning, medical knowledge service, as well as optimization of diagnosis and treatment process and improvement of management efficiency by combining various kinds of patient's diagnosis and treatment service data.

\textbf{Tongyi Renxin}: Tongyi Renxin is an AI product launched by Alibaba Cloud, which focuses on the medical field. It is based on the Tongyi generative LLM and integrates massive medical knowledge literature and medical data for training based on the total training data of Tongyi Qwen of more than 3 trillion tokens. Targeting the knowledge-intensive and serious characteristics of the pharmaceutical industry, the model has stronger industry knowledge reasoning and cognitive abilities in the medical field, including the accuracy of knowledge question answering, the authority of single round consultation, scenario-based multi-turn consultation, domain text generation, and domain literature understanding. At present, Tongyi Renxin is still in the invitation testing stage.

\subsection{Industrial Multimodal Medical LLMs}
In addition to text-based medical industrial LLM, the development of multimodal data has led to the emergence of multimodal industrial LLMs. Table \ref{tbl:in_multi_app} presents successful cases of multimodal medical LLM transitioning from academia to industry:

\textbf{StoneNeedle}: The Stone Needle LLM is the first multimodal LLM released by AthenaEyes that supports text, image, video, and audio inputs in the medical field and can provide an interactive experience that combines Chinese and Western medicine and integrates multimodality. Adopting the technical route of combining KG and LLM overcomes the pain point of low information accuracy of LLM technology in the medical field. At the same time, it can realize the multimodal processing of the text data of consultation, medical image data, video data of user's facial signs, and audio data of user's sleep to realize the diversified tasks such as medical auxiliary diagnosis, intelligent cognition, and health management. This revolutionizes the traditional single-task assisted diagnosis model in the healthcare industry.

\textbf{Tencent MedLLM}: Tencent MedLLM takes Tencent's newly released all-link self-research Hunyuan model as a base and continues to add medical KGs and medical literature covering 2.85 million medical entities, 12.5 million medical relationships, and 98\% of medical knowledge, to make the LLM further master professional medical knowledge. After 30 million Q\&A conversations covering patients, doctors, pharmaceutical companies, and other scenarios and medical processes for multi-task fine-tuning, as well as 360,000 sets of expert doctor-labeled data for reinforcement learning, it makes Tencent MedLLM more professional and precise in handling medical needs, while also taking into account patient care and getting closer to human doctors. It includes scenarios such as text generation, intelligent Q\&A, medical record structuring and retrieval, image reporting, and assisted diagnosis, and these all can be embedded in the whole process of medical sessions to achieve a comprehensive improvement in the level and quality of medical services.

\textbf{PanGu}: Based on the Pangu-Drug and relying on the Huawei Cloud Medical Intelligence Body EIHealth platform, the PanGu model has been trained with a super-large-scale model of compound characterization, and the chemical structures of 1.7 billion drug molecules have been pre-learned. It is mainly oriented to drug discovery and development, providing binding prediction, property prediction, molecular optimization, and generation capabilities. Pangu-Drug has a 20\% higher accuracy of drug-ability prediction than the traditional way, which helps researchers save a lot of the cost of drug design. In addition, the model has a built-in efficient molecular generator to generate a screening library of 100 million innovative drug-like small molecules with 99.68\% structural novelty, creating more possibilities for discovering new drugs. 

\textbf{Medical Sense}: Medical Sense, based on the large-scale language model "SenseChat" with 100 billion parameters developed by Shangtang, is trained with high-quality medical knowledge data of more than 20 billion tokens, which covers medical textbooks, medical guides, clinical paths, as well as 40 million real medical records, Q\&A, and dialogues between doctors and patients. Medical Sense focuses on four major application areas: intelligent health, intelligent patient service, intelligent clinic, and digital intelligence construction. It has covered 13 segmented medical and healthcare scenarios such as intelligent self-diagnosis, medical checkup consultation, and health Q\&A. Its goal is to accurately match model functions with specific scenarios and promote the digital transformation of the entire healthcare industry chain.

\textbf{MedLinker}: The MedLinker app is focused on real medical scenarios, using the MedGPT model with a 100 billion parameter scale. It is based on over 2 billion pieces of medical text data in the pre-training phase and 8 million high-quality structured clinical diagnosis and treatment data in the fine-tuning training phase. Over 100 doctors are involved in providing manual feedback to supervise the fine-tuning training. Currently, MedLinker can cover 60\% of ICD10 disease categories. Additionally, it overcomes the challenge of AI doctors not being able to engage in continuous free conversations with real patients for the first time. It supports multimodal input and output in medical consultation scenarios, fully realizing intelligence in disease prevention, diagnosis, treatment, and rehabilitation.

\textbf{WeiMai}: WeiMai and WeiMai-Doctor, based on Weimai's self-developed health management LLM model CareGPT, are mainly dedicated to giving full play to the value of health management in real healthcare service scenarios and realizing the full-cycle intelligent health management capabilities of prevention, consultation, appointment, and recovery. CareGPT combines a series of engineering tuning and full-course management technologies with a current parameter scale of 7 billion and can support medical health scenarios with multimodal inputs and outputs. Currently, WeiMai Total Disease Management covers 32 departments and more than 1,000 disease types, serving nearly 1 billion people.

\textbf{01Bot}: 01Bot, based on Baidu Wenxin, uses training corpus data of hundreds of billions of tokens, including massive clinical desensitization data, massive medical KGs, multimodal image data, health science contents, and clinical trial research information. It integrates the intelligent medical service experience of more than 800 hospitals and 4,000 grassroots clinics nationwide. It focuses on three major directions: intelligent health manager, intelligent physician assistant, and intelligent enterprise service, to satisfy the specific needs of "doctors, patients, and medicines." Intelligent Health Manager provides intelligent guidance, pre-questioning, and health consultation to provide patients with medical advice and guidance. Intelligent physician assistant provides services for physicians in terms of assisted diagnosis, medical record generation, and quick literature review. Intelligent enterprise services provide capabilities for enterprise customers, from operation assistance, vocational training, and knowledge services, helping enterprises quickly promote new drugs after market launch. 

\subsection{Academic vs Industrial}
The anchor models used in practical applications involve analyzing large datasets during the training process, which cover both public and non-public medical data. Moreover, many medical experts must participate in manual feedback supervision and fine-tuning training during the training process. Finally, to ensure the practicality and effectiveness of the model, a series of strict evaluation and validation measures need to be carried out before the model is applied.

Obviously, this is significantly different from academic medical LLMs in terms of training data, training process, and model evaluation. Subsequently, in the following section, we will compare industrial and academic medical LLM in more detail regarding research objectives, data resources, training methods, and application scenarios, as shown in Table \ref{tbl:ac_in}. 

\begin{table*}[h]
\footnotesize
    \centering
    \caption{Academic Medical LLMs versus Industrial Medical LLMs. We compare and analyze them with research objectives, data resources, training \& optimization, performance metrics, application scenarios, data privacy \& security, collaboration \& open source, and updates \& iterations.}
    \label{tbl:ac_in}
    \begin{tabularx}{\textwidth}{cXX}
    \hline
         & \textbf{Academic Medical LLMs} & \textbf{Industrial Medical LLMs}\\
         \hline
         \textbf{Research Objectives} & Research and innovation are often pursued to achieve theoretical breakthroughs and technological advancements. & While commercial interests may not be the primary driver, the ultimate goal is to solve practical problems and create commercial value in response to market demand. \\
         \hline
         \textbf{Data Resources} & Publicly available datasets are frequently used, or data is obtained through research collaborations with limited data volumes. & The organization utilizes a significant amount of data, including patient records and medical images, which are both diverse and voluminous. \\
         \hline
         \textbf{Training \& Optimization} & Focuses on innovations in model structure and algorithms, as well as theoretical performance improvements. & Focuses more on the practical application performance of the model, such as deployment efficiency, stability, and security. \\
         \hline
         \textbf{Performance Metrics} & Accuracy, Recall, F1 score, etc. & More diversified, including actual application effects, user feedback, business metrics, etc. \\
         \hline
         \textbf{Application Scenarios} & Demonstrate research results through academic research, published papers, and algorithmic competitions. & Focuses on the effectiveness of models in real clinical applications as well as commercial solutions. \\
         \hline
         \textbf{Data Privacy \& Security} & Often adhere to data privacy and security regulations, but may not be as strict compared to the industry. & Must adhere to strict healthcare data privacy and security regulations. \\
         \hline
         \textbf{Collaboration \&Open Source} & Prefers academic collaboration, paper publication, and open source code. & Prefers commercial collaboration and productization, technology licensing, may not be fully open source. \\
         \hline
         \textbf{Updates \& Iterations} & Relatively slow update frequency, mainly paper publication. & Iterate quickly to adapt to market needs and business changes. \\
         \hline
    \end{tabularx}
\end{table*}

Based on the differences between academic medical LLMs and industrial medical LLMs, we summarize the potential issues of implementing academic models into real-life applications in the following aspects:

\begin{itemize}
    \item \textbf{Limitations of Data Resources:} Academic research often utilizes publicly available datasets, while industry research relies more on private data resources that offer a wider range of real-world application scenarios and diverse cases. However, industry data may contain sensitive information that needs to be desensitized, potentially impacting the effectiveness of model training. Therefore, academic research may need to adjust to data quality and scale changes when transitioning to industry.
    \item \textbf{Balance of Model Performance and Practicality:} In academia, research models may prioritize theoretical innovation and performance improvement, whereas in industry, models must be stable and highly interpretable. As a result, industry application scenarios often require models with stronger generalization ability and robustness to handle various uncertainties in practical operations.
    \item \textbf{Clinical and Regulatory Compliance:} The healthcare industry requires models to meet strict clinical and ethical standards. Therefore, academic research results must undergo a rigorous clinical trial and regulatory approval when translated into industrial products.
    \item \textbf{Difficulty of Interdisciplinary Cooperation:} Differences in culture and working practices between academia and industry may impact interdisciplinary collaboration. Therefore, academic researchers may require a deeper understanding of the industry's needs and technical constraints, while industrial personnel may need to adapt to academic research methodologies.
    \item \textbf{Technology Transfer and Landing Challenges:} When applying academic research to practical industrial applications, it is often necessary to address specific issues such as hardware compatibility and computational resource limitations. Additionally, the industry may need to modify and engineer academic research findings before they can operate reliably in real clinical environments.
\end{itemize}

\section{Evaluation System}
\label{chp: 5}
In the fast-paced world of technology, evaluating models has become crucial for verifying their validity and reliability. A robust evaluation framework not only measures a model's performance but also guides its ongoing improvement and progression. This subsection will thoroughly examine the construction of the evaluation framework, covering its guiding principles, evaluation techniques, and real-world application scenarios to provide researchers with a comprehensive and up-to-date perspective.

\subsection{Assessment Principles}
The cornerstone of the evaluation system is a series of core principles that together ensure the assessment process's fairness, accuracy, and effectiveness. Let us take a closer look at these key principles:

\begin{itemize}
    \item \textbf{Accuracy}: The overarching principle of model assessment is accuracy, which refers to how closely the model predicts or generates results compared to the true or expected values. When evaluating mathematical questions, accuracy is determined by whether the model can correctly answer the questions. For language models, accuracy is reflected in the correctness and reasonableness of the content generated by the model.
    \item \textbf{Robustness}: The model can maintain efficient performance when faced with different data inputs, ensuring accuracy even in the presence of noise, missing data, or anomalies.
    \item \textbf{Generalization}: The model can maintain good performance when faced with unseen data, indicating a true understanding of the problem rather than just memorizing the training data.
    \item \textbf{Interpretability}: The workings and predictions of the model can be explained and understood, which is especially critical in scenarios involving important decisions.
    \item \textbf{Efficiency}: The time and resources consumed by the model in processing the task. In practice, efficient models can respond to requests faster and save computational resources.
    \item \textbf{Security}: Guaranteeing that the model is not maliciously exploited, ensuring data confidentiality during processing, and preventing the model's predictions from leading to undesirable consequences.
\end{itemize}

In addition to the above, other aspects, such as simplicity and novelty, can also be evaluated for the preferences of different models.

\subsection{Assessment Methodology}
Selecting appropriate assessment methods is essential to ensuring the quality and reliability of model evaluation results. When evaluating model performance, comprehensive considerations should be made to select assessment methods and metrics accurately reflecting the model's actual performance. This section will examine two primary assessment methods: automatic evaluation and human evaluation.

\subsubsection{Automatic Evaluation}
Automated quantitative metrics are used to assess model performance and applicability. This approach relies on algorithms and computational tools to quickly and objectively evaluate model performance. Its advantages are processing large amounts of data, reducing human bias, and allowing real-time updating of assessment results. Commonly automatic evaluation metrics such as accuracy, recall, and F1 score. They can reflect the model's predictive ability from different perspectives and provide strong support for quantitative analysis of the model. Here are some commonly automatic evaluation metrics:
\begin{itemize}
    \item \textbf{Accuracy} refers to the proportion of samples the model correctly predicts out of the total number of samples. The equation is as follows:
    \begin{equation} \label{acc}
    \begin{aligned}
        Accuracy=\frac{TP+TN}{TP+TN+FP+FN}
    \end{aligned}
    \end{equation}
    where TP stands for true positive, which means predicted positive and actual positive. FP stands for false positive, which means predicted positive but actual negative. FN stands for false negative, which means predicted negative but actual positive.
    \item \textbf{Precision} refers to the proportion of truly positive samples among those the model predicts as positive. The equation is as follows:
    \begin{equation} \label{pre}
    \begin{aligned}
        Precision=\frac{TP}{TP+FP}
    \end{aligned}
    \end{equation}
    where TP and FP have the same meaning as described in Equation \ref{acc}.
    \item \textbf{Recall} refers to the proportion of samples correctly predicted as positive by the model out of all the actual positive samples. The equation is as follows:
    \begin{equation} \label{rec}
    \begin{aligned}
        Recall=\frac{TP}{TP+FN}
    \end{aligned}
    \end{equation}
    where TP and FN have the same meaning as described in Equation \ref{acc}.
    \item \textbf{F1 score} is the harmonic mean of precision and recall, providing a comprehensive reflection of the model's accuracy and robustness. The equation is as follows:
    \begin{equation} \label{f1}
    \begin{aligned}
        F1 score=2\times \frac{Precision\times Recall}{Precision+Recall}
    \end{aligned}
    \end{equation}
    where Precision and Recall are described above.
    \item \textbf{Micro-F1\&Macro-F1} are F1 scores computed in macro-averaging and micro-averaging approaches, respectively, and are suitable for classification tasks. Micro-f1 aggregates the prediction results of all classes to compute global precision and recall. Macro-F1 calculates the F1 score for each class separately and then takes the average of all classes. In contrast, the latter is more suitable for multi-classification tasks.
    \item \textbf{BLEU} \cite{bleu} usually evaluates the quality of machine translation. It scores by comparing the overlap between the machine-generated translations and human translations.
    \item \textbf{ROUGE-L} \cite{rouge} assesses the performance of automatic summarization and machine translation. It calculates the longest common subsequence between the predicted and reference text.
    \item \textbf{METEOR} \cite{METEOR} also evaluates the quality of machine translation. It combines precision and recall, considering word matches and word order between the candidate and reference translation.
    \item \textbf{BERTScore} \cite{BERTScore}, based on the pre-trained language model BERT, evaluates the quality of text generation tasks such as machine translation and text summarization.
\end{itemize}

The above evaluation metrics, while commonly applied to a wide range of fields, their application in medicine needs to be appropriately adjusted to adapt to the uniqueness and complexity of medical tasks. For example, the medical concepts-based assessment metric, MEDCON \cite{Aci-bench}, is used to measure the accuracy and consistency of clinical concepts. This metric calculates the F1 score to determine the similarity between the UMLS concept set in the candidate and reference clinical notes. There is also the TCMScore metric \cite{TCMBench} for assessing TCM semantic and knowledge coherence, which combines the matching of TCM terms and semantic consistency between the generated and standard analysis. In this case, term matching calculates by adding term diversity to the original F1 score calculated from precision and recall, thus evolving into a Term F1 Score.

\subsubsection{Human Evaluation}
Model's qualitative analysis and evaluation outputs by professionals. Compared with automatic evaluation, human evaluation focuses on subjectively evaluating the model outputs using the knowledge and experience of the experts, especially where the model is in areas that are difficult to capture in automatic evaluation, such as the interpretability and robustness of the model. In addition, this approach also focuses on the utility and user acceptance of the model. Human evaluation typically includes user testing, expert reviews, and case studies, which can provide insights into the qualitative analysis of the model.

Currently, medical academic LLMs emphasize the model's performance in response to specific medical scenarios, such as medical exams and QA. These tasks are often derived from real medical data and evaluated for accuracy using an automatic matching method. Common evaluation datasets include MultiMedQA, MultiMedBench, CMB \cite{CMB}, PromptCBLUE \cite{PromptCBLUE}, RJUA-QA Datasets \cite{RJUA-QA}, EHRNoteQA \cite{EHRNoteQA}, Aci-bench \cite{Aci-bench} and TCM-ED \cite{TCMBench}. 
While these evaluation datasets provide diverse medical information, including images, text, and clinical Q\&A, and some also introduce human evaluation frameworks to improve data accuracy, they have some limitations. For example, MultiMedQA has limited labeled data based on diagnostic reports and can only accurately assess a limited number of tasks. MultiMedBench has a limited amount of data in the transcriptomics and proteomics domains. CMB, PromptCBLUE, RJUA-QA Datasets, and TCM-ED still have limitations in terms of data diversity, and the EHRNoteQA and Aci-bench are insufficient in task diversity. They all have room for improvement in data scale or disease coverage.

\subsection{Summary and Discussion}
\begin{figure}[!h]
    \centering
    \includegraphics[width=1\textwidth]{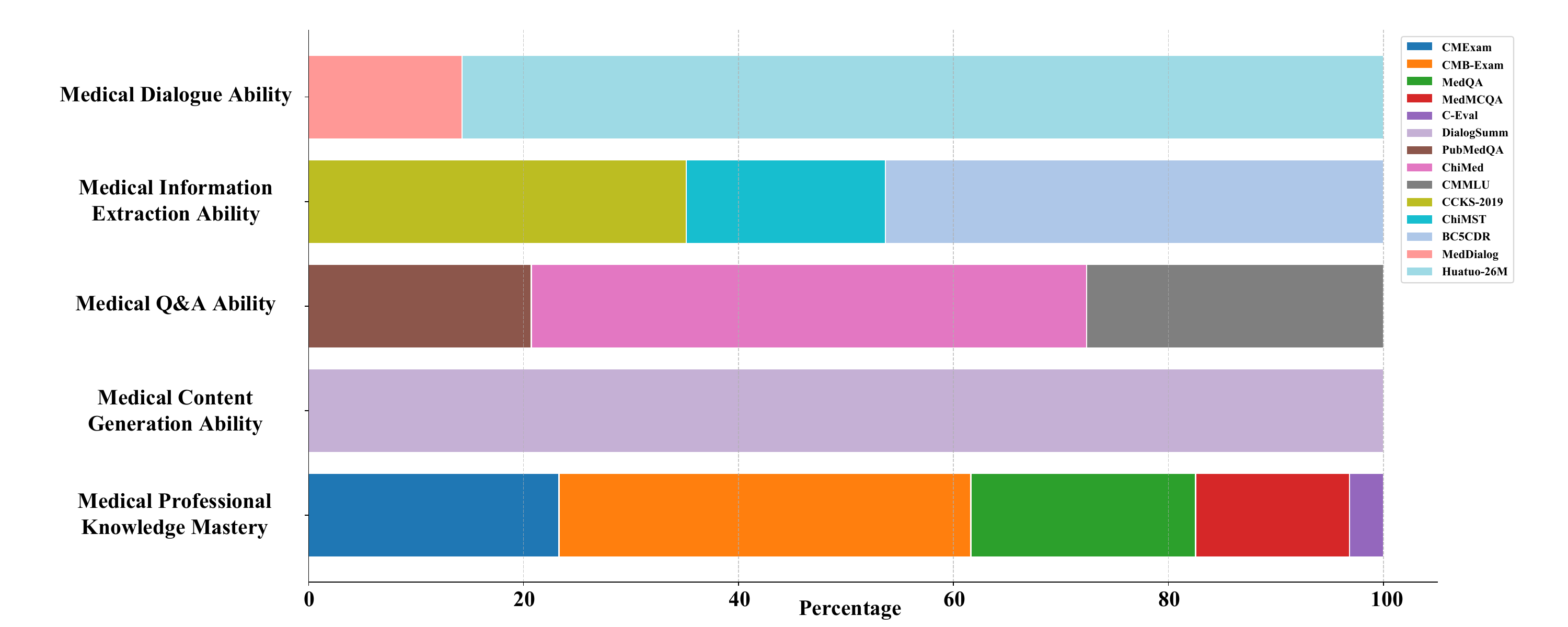}
    \caption{Data distribution of different capabilities of text medical LLMs.}
    \label{fig:data-distribution-text}
\end{figure}

\begin{figure}[!htp]
    \centering
    \includegraphics[width=1\textwidth]{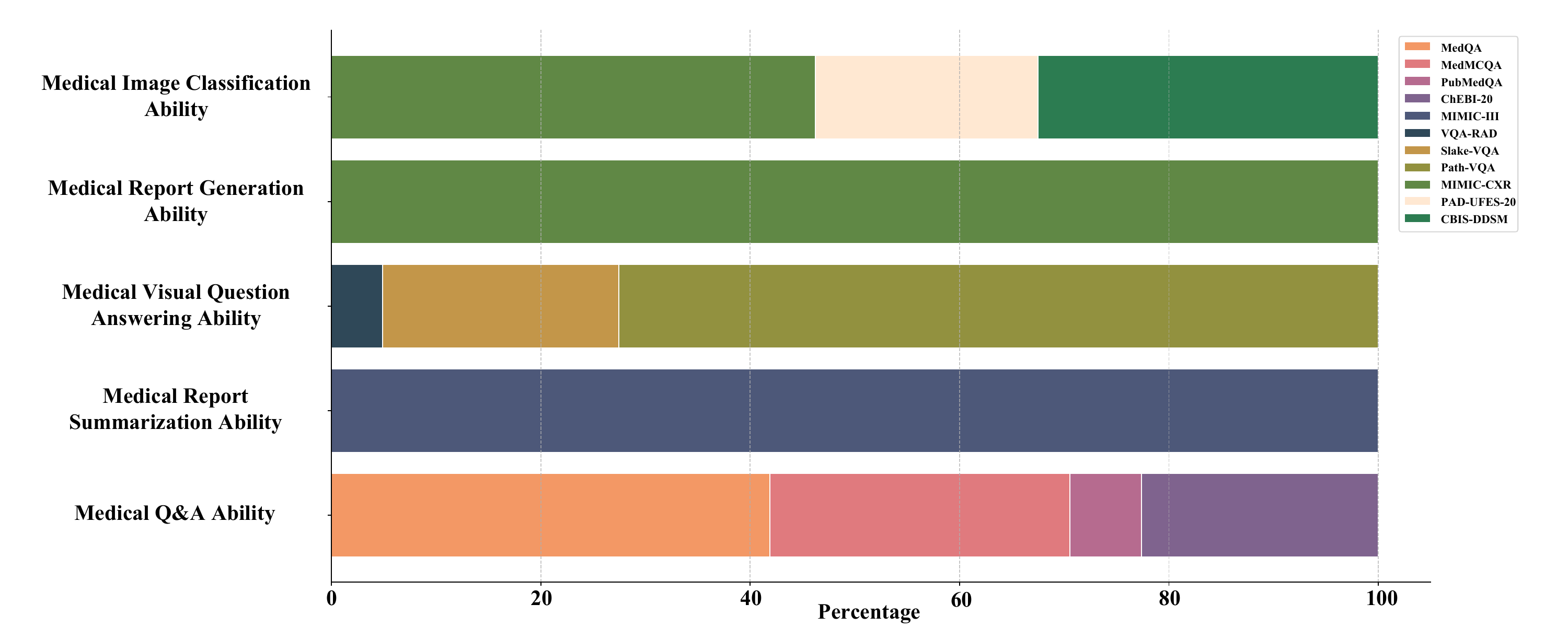}
    \caption{Data distribution of different capabilities of multimodal medical LLMs.}
    \label{fig:data-distribution-multimodal}
\end{figure}

We systematically summarize the data distribution of academic LLMs across different medical capabilities (see Figures \ref{fig:data-distribution-text} and \ref{fig:data-distribution-multimodal} for details) and conduct a comparative analysis of these models' performance in medical capabilities (see \ref{t-llms} and \ref{m-llms} for details). Our research finds that there are significant differences between models when assessing the same medical capabilities, mainly in the following aspects:
\begin{itemize}
    \item \textbf{Dataset Inconsistency.} Some models are evaluated using different datasets, making direct comparisons difficult.
    \item \textbf{Variability in Evaluation Methods and Metrics.} Most models employ automated evaluation methods. However, the evaluation metrics were not entirely uniform, even within the same dataset.
    \item \textbf{Limitations of Human Evaluation.} A few models incorporate human evaluations, but their evaluation systems vary, hindering unified assessments across models.
\end{itemize}

These issues limit fair and systematic comparisons between models with the same capabilities and affect the generalizability and reproducibility of the research findings.

Although automatic evaluation methods can provide an initial performance quantification, their limitation lies in the difficulty in capturing the complexities and nuances of medical clinical practice. In contrast, human evaluation enables medical experts to analyze model outputs in depth, especially in key areas such as diagnostic recommendations, treatment recommendations, and patient management strategies. Its importance is not only for its ability to identify details missed by automatic evaluation but also for validating the clinical adaptability of the model's decision-making process. Furthermore, human evaluation helps to identify and correct model biases, enhancing the fairness and ethicality of the model, while feedback from medical experts is crucial for the continuous training and optimization of the model.

Given the specificities of the medical field, we recommend a combined approach of both automatic and human evaluations when assessing models. Integrating the efficiency of automatic evaluation with the depth of human evaluation, we can create a comprehensive assessment framework. This framework enhances the thoroughness and depth of model evaluations while ensuring the reliability and effectiveness of models in practical applications. Specifically, future research should focus on the following points:
\begin{enumerate}
    \item \textbf{Establish Unified Evaluation Benchmarks and Datasets.} This provides a foundation for horizontal comparisons between different models.
    \item \textbf{Clarify Evaluation Metrics and Methods.} This will improve the comparability and consistency of evaluation results.
    \item \textbf{Develop a Standardized Human Evaluation System.} This aims to reduce subjective bias in the evaluation process.
\end{enumerate}

Through comprehensive evaluations, we can gain a more holistic understanding of model performance, identify their strengths and limitations, and ultimately provide better services to patients while guiding further development and application of the models. This approach promotes the application of LLMs in the medical field and enhances the transparency and credibility of research, thereby adding value to medical research and practice.

\section{Challenges and Future Work}
\label{chp: 6}
\subsection{Challenges}
Medical LLMs are gaining academic researcher's attention due to their advanced language understanding and generation capabilities. The development of academic medical AI systems has enabled models to process and generate complex medical texts, making them useful for tasks such as medical QA, medical dialogue, medical text generation, and medical knowledge graph construction.

However, despite the exciting research results of medical LLMs, there are still challenges in their academic research and practical implementation. As far as academic research is concerned, there are many challenges in the whole process, from data to models and evaluation.

\begin{itemize}
    \item \textbf{Clinical Databases and Datasets:} The quantity, quality, and accuracy of existing medical knowledge bases and training datasets need to be expanded and improved due to medical data's high complexity and diversity. Furthermore, multi-language support is still lacking. Additionally, the disease knowledge in these knowledge bases and datasets may not be comprehensive enough. For instance, the information on some rare or endemic diseases is not detailed enough, and the latest research results and treatments are not updated in time.
    \item \textbf{Model Construction:} Although LLM model construction based on medical text corpora or multimodal data shows excellent performance in some or more medical tasks, it still faces several problems. These include data annotation quality, uneven data distribution, lack of model interpretability and transparency, and limitations in multimodal information fusion.
    \item \textbf{Assessment Systems:} During the assessment of model performance, most methods rely on automated medical task evaluation metrics. They generally focus on specific medical tasks, such as medical examinations and QA. However, relatively few methods incorporate manual assessment. In particular, many evaluation datasets are deficient in data size or disease coverage. This can compromise the overall assessment of model performance.
    \item \textbf{Collaborative knowledge graphs:} Combining KG with LLM can enhance the model's NLP abilities while ensuring the medical field's interpretability, trustworthiness, and traceability. However, challenges such as catastrophic forgetting and erroneous knowledge editing when updating LLM knowledge with knowledge graphs or LLM's illusions still need to be addressed. Furthermore, additional research is required to investigate effective techniques for integrating knowledge into LLM using KGs.
\end{itemize}

Despite these technical concerns for developing academic medical LLMs, we still need to tackle the following endeavors while putting these models into real-life applications:

\begin{itemize}
    \item \textbf{Evaluations:} As a common evaluation strategy for evaluating model performances, accuracy usually refers to the rate of correct predictions by a model. In clinical applications, doctors often focus on the number of specific information that a model can correctly identify or provide. In this case, even if a model has a high accuracy rate, it may still fail to satisfy the clinical decision requirements such as drug dosage and treatment duration. These high-quality models only have a higher probability of producing correct answers, which will increase the skepticism. Therefore, more practical evaluation metrics with better reflections of model capacities in real-life practice are needed.
    \item \textbf{Diversity:} Academic research findings may only be applicable under specific conditions, while practical applications must consider a wider patient population, different etiologies, and comorbidities. 
    \item \textbf{Interpretability:} Additionally, models must be interpretable, traceable, and evidence-based to gain the trust of both doctors and patients and to be effectively applied in real clinical environments for the convenience of doctors and patients.
\end{itemize}

In addition to the challenges mentioned above, data privacy, data security, and the corresponding regulatory and ethical issues are equally important for medical LLM practices \cite{Jasmine24Medical}. These concerns must be strictly adhered to, whether in academic research or practical application.

\subsection{Future Work}
To address these challenges, further investigations can consider the following paths:
\begin{enumerate}
    \item In academic research, improving the medical knowledge base \cite{Yang24LMKG, Chandak23Building} can enhance the accuracy of the models. Similarly, larger and more diverse evaluation datasets \cite{Sandeep23Evaluating} can help in the multi-dimensional assessment of models and validate their clinical applicability.
    \item Increasing the transparency \cite{Maharjan24OpenMedLM} of the models can improve their interpretability \cite{Savage2024Diagnostic}. Additionally, closely integrating the models with clinical practice can ensure their traceability and evidence-based transferability \cite{Tang23MedAgents, Tang23Evaluating}.
\end{enumerate}

With the accumulation of medical data and optimization of algorithms, medical LLMs are expected to perform even better in real-world applications. This will provide doctors with more accurate decision-making assistance and patients with higher-quality medical services.

Other than technical issues, mitigating the doctor-patient relationship and providing humanistic care can also be integrated into academic medical LLMs research:
\begin{enumerate}
    \item \textbf{Optimising Patients' Medical Experience:} By analyzing patients' behavioral patterns and combining them with the construction of a doctor-patient trust model, we can predict and enhance trust between doctors and patients through LLMs. This will optimize the medical service process and reduce patients' waiting time. Furthermore, we can offer patients customized and accurate medical services, along with personalized health education and programs for managing their conditions. These improvements will enhance patient satisfaction and treatment compliance.  
    \item \textbf{Medical Resources and Service Improvement:} We can utilize LLMs for the optimal allocation of healthcare resources to ensure that resources are fairly and reasonably distributed across various regions and groups. Meanwhile, we can use the LLMs to analyze the causes and patterns of medical disputes, propose preventive and handling strategies, reduce doctor-patient conflicts, ensure medical safety, and enhance the quality of healthcare services.
    \item \textbf{Patient Psychological Support and Cross-Cultural Communication:} We can employ LLMs to provide patients with mental health assessments and support, identify psychological issues, and offer psychological therapies and interventions \cite{Wang24Apollo}. Additionally, we can use the LLMs to achieve cross-cultural doctor-patient communication, providing more considerate and understanding medical services for patients from diverse cultural backgrounds \cite{Pieri24BiMediX}. This will reduce misunderstandings and conflicts caused by cultural differences, thereby enhancing patients' mental health and overall healthcare experience.
\end{enumerate}

\section{Conculsion}
Clinical knowledge about the causes, prognosis, diagnosis, and treatment of diseases can improve curing performances, and promote physical health. The recent development of LLMs opens new possibilities in the field of medical AI. In this survey, we gather the building paradigms of medical AI systems including the use of clinical databases, datasets, training pipelines, evaluation systems, as well as methods integrating medical KGs. 

Finally, we present the differences between academic medical LLMs and industrial ones and summarize the challenges to implementing academic LLMs in real-life medical situations. Based on this analysis, we present some of the future directions for academic research and applications. We hope that our survey presents an overview of the field of medical AI and its applications as well as challenges and future works for these technologies.


\appendix
\section{Industrial Text Medical LLMs}
Here we present the summarization of industrial medical LLMs building from text corpora.

\begin{sidewaystable}[!htp]
\footnotesize
\begin{threeparttable}
	\caption{Application for plain-text industrial medical LLMs. These models often were built on vast real-life clinical data collected by cooperation's, which reflects the patient condition more comprehensively.}
        \label{tbl:in_cor_app}
	\centering
	\begin{tabular}{ccccc}
        \hline
		\textbf{Application Name} & \textbf{Anchor Model} & \textbf{Data Scale} & \textbf{Enterprise Name} & \textbf{Function} \\
            \hline
		UniGPT-Med & UniGPT \tnote{1} & 2,000B tokens & Unisound AI Technology Co., Ltd. & \makecell{AI-assisted D\&T\\Medical Record Generation}\\
            \hline
            - & EyeGPT \tnote{2} & - & Wenzhou International Optometry Innovation Centre & Clinical Medical Assistance \\
            \hline
             DaJing TCM & QiHuangWenDao\tnote{3} & \makecell{11M TCM-KG\\1,500 books\&literatures\\100,000 real medical cases\\100,000 P\&T\&M\&A\tnote{4}\\2M diagnosis} & Nanjing Dajing TCM IT Co., Ltd. & \makecell{Intelligent Assisted D\&T\\Intelligent Health Conditioning} \\
            \hline
            DingDangKuaiYao\tnote{5} & HealthGPT & - & DingDangKuaiYao Technology Group Co., Ltd. & \makecell{Quick Medicine Service\\Online Consultation\\Chronic and Health Management}\\
            \hline
            Gushengtang TCM & \multirow{2}{*}{GushengtangTCM} & \multirow{2}{*}{-} & \multirow{2}{*}{Gushengtang TCM Chain Management Group} & \multirow{2}{*}{\makecell{AI-assisted D\&T\\Recommended TCM Formulas}}\\
            Gushengtang-Doctor & & & & \\
            \hline
            Xunfei Xiaoyi & Xinghuo\tnote{6} & - & iFLYTEK Co., Ltd. & \makecell{Report Interpretation\\TCM Syndromes Identification} \\
           \hline
            Zhiyun Health\tnote{7} & ClouD GPT & - & Hangzhou Comms IT Co., Ltd. & \makecell{AI-assisted Diagnosis\\AI Drug\\Device Development}\\
            \hline
            WiNEX\tnote{8} & WiNGPT\tnote{9} & 30B/65B tokens & Winning Health Technology Group Co.,Ltd & Health Care Assistant\\
            \hline
            Tongyi Renxin\tnote{10} & Tongyi Qwen\tnote{11} & $>$3,000B tokens & Alibaba Group & \makecell{Intelligent Inquiry\\Report Interpretation\\Summary Abstract} \\
        \hline
	\end{tabular}
        \begin{tablenotes}
		\item[1] http://shanhai.unisound.com/
            \item[2] http://eyegpt.com.cn/\#/
            \item[3] http://www.dajingtcm.com/dajinggpt
            \item[4] Pulse\&Tongue\&Meridian\&Acupuncture
            \item[5] https://www.ddky.com/
            \item[6] https://xinghuo.xfyun.cn/
            \item[7] https://www.zyhealth.com
            \item[8] https://www.winning.com.cn/WiNEX/
            \item[9] https://github.com/winninghealth/WiNGPT2
            \item[10] https://tongyi.aliyun.com/renxin
            \item[11] https://tongyi.aliyun.com/qianwen
        \end{tablenotes}
 \end{threeparttable}
\end{sidewaystable}

\section{Industrial Multimodal Medical LLMs}
Here we present the summarization of industrial medical LLMs building from multimodal data.

\begin{sidewaystable}[!h]
\footnotesize
\begin{threeparttable}
	\caption{Application for multimodal industrial medical LLMs. The integration of numerous clinical data gives the model a better understanding over the fine-grained medical knowledge.}
        \label{tbl:in_multi_app}
	\centering
	\begin{tabular}{ccccc}
        \hline
		\textbf{Application Name} & \textbf{Anchor Model} & \textbf{Data Scale} & \textbf{Enterprise Name} & \textbf{Function} \\
            \hline
		StoneNeedle\cite{StoneNeedle} & - & - & AthenaEyesCo., LTD. & \makecell{AI-assisted Diagnosis\\AI-assisted Reading\\Physiological Prediction\\Sleep Monitoring}\\
            \hline
            Tencent MedLLM & Hunyuan\tnote{1} & \makecell{2.85M medical entities\\12.5M medical relations\\Medical KG\&literatures covering 98\%\\30M dialogues\\360,000 expert’s annotation} & Tencent Cloud Computing (Beijing) Co., Ltd. & \makecell{Content Generation\\Medical Record Structuralization\\AI-assisted Diagnosis\\AI Rational Drug Use\\AI-assisted Reading}\\
            \hline
            PanGu & Pangu-Drug\cite{Weihua23Stone} & $>$3,000B tokens & Huawei Technologies Co., Ltd. & AI-assisted drug R\&D \\
            \hline
            Medical Sense & SenseChat\tnote{2} & $>$20B tokens & Shanghai SenseTime IT Co., Ltd. & \makecell{AI-assisted diagnosis\\Clinic Interpreter Robot\\Medical Record Structuralization} \\
            \hline
            MedLinker\tnote{3} & MedGPT & \makecell{$>$2B medical texts\\SFT 8M diagnosis\\$>$100 doctors RLHF} & Chengdu MedCloud Technology Co., Ltd. & Intelligent Health Inquiry\\
            \hline
            WeiMai & \multirow{2}{*}{CareGPT} & \makecell{$>$1B medical texts\\Millions of medical and health KB} & \multirow{2}{*}{Weima Technology Co., Ltd.} & \multirow{2}{*}{\makecell{Personalized Matching and Recommendation\\Aiding Decision-making}}\\
            WeiMai-Doctor & & SFT $>$100 doctors RLHF & & \\
           \hline
            01Bot\tnote{4} & Baidu Wenxin\tnote{5} & Hundred billion tokens & Baidu Online Network Technology (Beijing) Co., Ltd. & \makecell{AI-assisted Diagnosis\\Medical Record Generation}\\
        \hline
	\end{tabular}
        \begin{tablenotes}
		  \item[1] https://hunyuan.tencent.com/
            \item[2] https://chat.sensetime.com/
            \item[3] https://www.medlinker.com/
            \item[4] https://01.baidu.com/bot.html
            \item[5] https://wenxin.baidu.com/
        \end{tablenotes}
 \end{threeparttable}
\end{sidewaystable}

\section{Academic Medical Texts LLMs Assessment}
\label{t-llms}
Here, we systematically summarize the performance of academic medical texts LLMs across five key medical capabilities: medical professional knowledge mastery, medical Q\&A ability, medical information extraction ability, medical dialogue ability, and medical content generation ability. Moreover, we provide a detailed overview of their evaluation methods.

\begin{sidewaystable}[!htp]
\tiny
\begin{threeparttable}
    \caption{Summary of academic medical texts LLMs in medical professional knowledge mastery.}
    \label{tab:t-llms1}
    \centering
        \begin{tabular}{ccccccc}
        \hline
        \multirow{4}{*}{} & \multirow{4}{*}{Assessment} & \multicolumn{5}{c}{\textbf{Medical Professional Knowledge Mastery}} \\
        & & CMExam & CMB-Exam & MedQA & MedMCQA & C-Eval \\
        & & CMExam Prediction $\mid$ CMExam Reasoning & Physician $\mid$ Nurse $\mid$ Pharmacist $\mid$ Technician $\mid$ Disciplines $\mid$ Graduate Entrance & USMLE $\mid$ MCMLE $\mid$ TWMLE & & Clinical Medicine $\mid$ Physician $\mid$ Basic Medicine \\
        & & Acc $\mid$ BLEU-1/4$\big/$Rouge-1/2/L & Acc Avg. & Acc $\mid$ Elo Rating & Acc & Acc \\
        \hline
        BioMedLM-2.7B & Auto & & & 50.3// $\mid$ & 57.3 & \\
        GatorTronGPT-5B & Auto$+$Human & & & 40.2// $\mid$ & 35.8 & \\
        GatorTronGPT-20B & Auto$+$Human & & & 45.1// $\mid$ & 42.9 & \\
        PMC-LLaMA-13B & Auto & & & 56.36// $\mid$ & 56.04 & \\
        Med-PaLM & Auto$+$Human & & & 67.2// $\mid$ & & \\
        Med-PaLM 2(Ensemble Refinement) & Auto$+$Human & & & 85.4// $\mid$ & 72.3 & \\
        Med-PaLM 2 & Auto$+$Human & & & 86.5// $\mid$ & 72.3 & \\
        ChatDoctor-7B & Auto & & & 33.93// $\mid$ & 31.1 & \\
        \hline
        Sunsimiao(Baichuan) & Auto & & 38.75/44.37/38.81/38.33/37.5/33.31 & & & \\
        Sunsimiao-7B(Qwen) & Auto & & 66.75/72/71.53/64.83/63.06/75.19 & & & \\
        QiZhenGPT-QiZhen-CaMA-13B-Checkpoint-12400 & CS(Drug Indication Review/Disease Review) & & & $\mid$ 955/959/ & & \\
        ChatMed-Consult & CS(Patients' Daily Medical Problems) & & 21.41/23.48/21.58/23.55/21.36/18.08 & & & \\
        BenTsao & Human & 12.9 $\mid$ 0.21/0.12/25.11/11.56/9.73 & 21.67/19.99/21.07/22.85/19.83/16.93 & $\mid$ 961/921/ & & \\
        ClinicalGPT & Auto$+$GPT4 & & & /37.6\tnote{2}/ $\mid$ & & \\
        MedicalGPT(Baichuan) & CS(Medical Q\&A and Daily Q\&A) & & & /18.5/ $\mid$ & & 39.02\tnote{3} \\
        MedicalGPT(Ziya) & CS(Medical Q\&A and Daily Q\&A) & & 26.56/30.94/24.72/27.17/25.44/21.5 & /25.99/ $\mid$ & & 48.78\tnote{3} \\
        HuatuoGPT & Auto$+$GPT4$+$Human & 31.07 $\mid$ & 31.85/35/30.56/31.38/35/28.25 & 25.77// $\mid$ & 31.2 & 36.53\tnote{3} \\
        HuatuoGPT2\tnote{1} & - & & & $\mid$ 955/993/ & & \\
        HuatuoGPTII-7B & Auto$+$GPT4$+$Human & 65.81 $\mid$ & 64.55/63.75/61.06/56.25/56.63/51.81 & 41.13// $\mid$ & 41.87 & 62.4\tnote{3} \\
        HuatuoGPTII-13B & - & 68.98 $\mid$ & 62.75/66.13/64.91/62/61.94/53.69 & 45.68// $\mid$ & 47.41 & 64\tnote{3} \\
        HuatuoGPTII-34B & - & & 75.65/82.31/76.81/76.17/74.38/75.56 & & & \\
        DISC-MedLLM & Auto$+$GPT3.5/GPT4 & 36.62 $\mid$ & 42.25/46.88/38.44/38.83/40.75/31.44 & 28.67// $\mid$ & - & - \\
        BianQue & Auto & & & & & \\
        BianQue 2 & Auto & & 6.95/7.31/7.25/9.75/6.94/6.06 & $\mid$ 913/928/ & & \\
        DoctorGLM & Auto$+$Semi-human & - $\mid$ 9.43/2.65/21.11/6.86/9.99 & 6.95/7.31/8.28/9.75/7.5/6.06 & $\mid$ 906/896/ & & \\
        PULSE-20b(InternLM) & Auto & & & $\mid$ 1042/1024/ & & \\
        ChiMed-GPT & Auto$+$Human & & & /44.5/ $\mid$ & & 68.29\tnote{3} \\
        Qilin-Med-7B-CPT & Auto & 38.4 $\mid$ 13.98/4.43/23.51/8.68/7.41 & & & & 41/44.9/34.3 \\
        Qilin-Med-7B-SFT & Auto & 40 $\mid$ 40.31/25.05/53.56/36.39/34.17 & & & & 48.5/55.5/43.4 \\
        Qilin-Med-7B-DPO & Auto & & & & & \\
        \hline
        \end{tabular}
    \begin{tablenotes}
        \item[1] https://www.huatuogpt.cn/\#/
         \item[2] Respiratory-Urinary-Digestive-Rheumatic immune
        \item[3] $(Clinical\ Medicine+Basic\ Medicine)/2$
    \end{tablenotes}
\end{threeparttable}
\end{sidewaystable}

\begin{sidewaystable}[!htp]
\tiny
\begin{threeparttable}
    \caption{Summary of academic medical texts LLMs in medical Q\&A and information extraction ability.}
    \label{tab:t-llms2}
    \centering
        \begin{tabular}{cccccccc}
        \hline
        \multirow{4}{*}{} & \multicolumn{4}{c}{\textbf{Medical Q\&A Ability}} & \multicolumn{3}{c}{\textbf{Medical Information Extraction Ability}} \\
        & PubMedQA & ChiMed & CMMLU & iCliniq & CCKS-2019 & ChiMST & BC5CDR \\
        & & & Anatomy $\mid$ Clinical Knowledge $\mid$ College Medicine $\mid$ Genetics $\mid$ Nutrition $\mid$ Traditional Chinese Medicine $\mid$ Virology & & & & \\
        & Acc & BLEU-1/2$\big/$Rouge-1/2/L & Acc & Precision$\big/$Recall$\big/$F1 & F1 & F1 & Precision$\big/$Recall$\big/$F1 \\
        \hline
        BioMedLM-2.7B & 74.4 & & & & & & \\
        GatorTronGPT-5B & 75.8 & & & & & & 58.7/43.4/47.2 \\
        GatorTronGPT-20B & 77.6 & & & & & & 54.3/49.9/49.4 \\
        PMC-LLaMA-13B & 77.9 & & & & & & \\
        Med-PaLM & & & & & & & \\
        Med-PaLM 2(Ensemble Refinement) & 75 & & & & & & \\
        Med-PaLM 2 & 81.8 & & & & & & \\
        ChatDoctor-7B & 54.3 & & & 84.44$\pm$1.85/84.51$\pm$1.57/84.46$\pm$1.38 & & & \\
        \hline
        Sunsimiao(Baichuan) & & & & & & & \\
        Sunsimiao-7B(Qwen) & & & & & & & \\
        QiZhenGPT-QiZhen-CaMA-13B-Checkpoint-12400 & & & & & & & \\
        ChatMed-Consult & & & & & & & \\
        BenTsao & & & & & & & \\
        ClinicalGPT & & & & & & & \\
        MedicalGPT(Baichuan) & & 5.82/5.26/16.61/2.94/11.11 & 43.82 & & 23.8 & 26.16 & \\
        MedicalGPT(Ziya) & & 39.02/32.35/26.76/8.10/18.16 & 34.56 & & 29.59 & 28.12 & \\
        HuatuoGPT & & & 33.23 & & & & \\
        HuatuoGPT2\tnote{1} & & & & & & & \\
        HuatuoGPTII-7B & & & 59.08 & & & & \\
        HuatuoGPTII-13B & & & 61.45 & & & & \\
        HuatuoGPTII-34B & & & & & & & \\
        DISC-MedLLM & & & - & & & & \\
        BianQue & & & & & & & \\
        BianQue 2 & & & & & & & \\
        DoctorGLM & & & & & & & \\
        PULSE-20b(InternLM) & & & & & & & \\
        ChiMed-GPT & & 44.58/37.22/27.11/8.89/19.86 & 52.92 & & 40.82 & 41.04 & \\
        Qilin-Med-7B-CPT & & & & & & & \\
        Qilin-Med-7B-SFT & & & & & & & \\
        Qilin-Med-7B-DPO & & & & & & & \\
        \hline
        \end{tabular}
    \begin{tablenotes}
        \item[1] https://www.huatuogpt.cn/\#/
    \end{tablenotes}
\end{threeparttable}
\end{sidewaystable}

\begin{sidewaystable}[!htp]
\footnotesize
\begin{threeparttable}
    \caption{Summary of academic medical texts LLMs in medical dialogue and content generation ability.}
    \label{tab:t-llms3}
    \centering
        \begin{tabular}{cccc}
        \hline
        \multirow{4}{*}{} & \multicolumn{2}{c}{\textbf{Medical Dialogue Ability}} & \textbf{Medical Content Generation Ability} \\
        & MedDialog-CN & Huatuo-26M & DialogSumm \\
        & & & \\
        & BLEU-1/2/3/4$\big/$Rouge-1/2/L & BLEU-1/4$\big/$Rouge-1/2/L & Elo Rating \\
        \hline
        BioMedLM-2.7B & & & \\
        GatorTronGPT-5B & & & \\
        GatorTronGPT-20B & & & \\
        PMC-LLaMA-13B & & & \\
        Med-PaLM & & & \\
        Med-PaLM 2(Ensemble Refinement) & & & \\
        Med-PaLM 2 & & & \\
        ChatDoctor-7B & & & \\
        \hline
        Sunsimiao(Baichuan) & & & \\
        Sunsimiao-7B(Qwen) & & & \\
        QiZhenGPT-QiZhen-CaMA-13B-Checkpoint-12400 & & & 921 \\
        ChatMed-Consult & & & \\
        BenTsao & & & 920 \\
        ClinicalGPT & 13.9/3.7/2.0/1.2/27.9/6.5/21.3 & & \\
        MedicalGPT(Baichuan) & & & \\
        MedicalGPT(Ziya) & & & \\
        HuatuoGPT & & 25.16/4.4/27.76/7.45/17.99 & \\
        HuatuoGPT2\tnote{1} & & & 980 \\
        HuatuoGPTII-7B & & & \\
        HuatuoGPTII-13B & & & \\
        HuatuoGPTII-34B & & & \\
        DISC-MedLLM & & & \\
        BianQue & 11.12/6.5/4.42/3.1/15.55/2.15/12.96 & & \\
        BianQue 2 & & & 908 \\
        DoctorGLM & 10.39/5.06/2.94/1.8/13.27/1.04/11.17 & & 905 \\
        PULSE-20b(InternLM) & & & 1076 \\
        ChiMed-GPT & & & \\
        Qilin-Med-7B-CPT & & 10.63/0.98/19.97/3.33/4.94 & \\
        Qilin-Med-7B-SFT & & 12.69/2.07/24.21/6.34/11.56 & \\
        Qilin-Med-7B-DPO & & 16.66/2.64/27.44/6.88/9.36 & \\
        \hline
        \end{tabular}
    \begin{tablenotes}
        \item[1] https://www.huatuogpt.cn/\#/
    \end{tablenotes}
\end{threeparttable}
\end{sidewaystable}

\section{Academic Medical Multimodal LLMs Assessment}
\label{m-llms}
Here, we systematically summarize the performance of academic medical multimodal LLMs across five key medical capabilities: medical Q\&A ability, medical report summarization ability, medical visual question answering ability, medical report generation ability, and medical image classification ability. Additionally, we provide a detailed overview of their evaluation methods.

\begin{sidewaystable}[!htp]
\footnotesize
\begin{threeparttable}
    \caption{Summary of academic medical multimodal LLMs in medical Q\&A ability.}
    \label{tab:m-llms1}
    \centering
        \begin{tabular}{ccccccc}
        \hline
        \multirow{4}{*}{} & \multirow{4}{*}{Assessment} & \multicolumn{5}{c}{\textbf{Medical Q\&A Ability}} \\
        & & MedQA & MedMCQA & PubMedQA & ChEBI-20 & UniProtQA \\
        & & USMLE $\mid$ MCMLE $\mid$ TWMLE & & & & \\
        & & Acc & Acc & Acc & BLEU-2/4$\big/$Rouge-1/2/L$\big/$METEOR & BLEU-2/4$\big/$Rouge-1/2/L$\big/$METEOR \\
        \hline
        BioMedGPT-10B & Auto & 50.4// & 51.4 & 76.1 & 23.4/14.1/38.6/20.6/33.2/30.8 & 57.1/53.5/74.3/75.9/62.2/75.4 \\
        Med-PaLM M-12B & Auto$+$Human & 29.22// & 32.2 & 48.6 & & \\
        Med-PaLM M-84B & Auto$+$Human & 46.11// & 47.6 & 71.4 & & \\
        Med-PaLM M-562B & Auto$+$Human & 69.68// & 62.59 & 80 & & \\
        Visual Med-Alpaca & CS(ROCO) & & & & & \\
        BiomedGPT\tnote{2}-S-33M & Auto$+$Human & & & & & \\
        BiomedGPT-M-93M & Auto$+$Human & & & & & \\
        BiomedGPT-B-182M & Auto$+$Human & & & & & \\
        \hline
        Qilin-Med-VL & CS(PMC-VQA) & & & & & \\
        LLaVA-Med(BioMed CLIP) & Auto & & & & & \\
        XrayPULSE & CS(MIMIC-CXR/OpenI) & & & & & \\
        XrayGLM & CS(MIMIC-CXR/OpenI) & & & & & \\
        \hline
        \end{tabular}
    \begin{tablenotes}
        \item[1] Where ``Auto'' means ``Automatic Evaluation''. ``Human'' means ``Human Evaluation''. ``CS'' means ``Case Study''.
        \item[2] https://github.com/taokz/BiomedGPT
    \end{tablenotes}
\end{threeparttable}
\end{sidewaystable}

\begin{sidewaystable}[!htp]
\footnotesize
\begin{threeparttable}
    \caption{Summary of academic medical multimodal LLMs in medical report summarization and visual question answering ability.}
    \label{tab:m-llms2}
    \centering
        \begin{tabular}{ccccccccc}
        \hline
        \multirow{4}{*}{} & \textbf{Medical Report Summarization Ability} & \multicolumn{3}{c}{\textbf{Medical Visual Question Answering Ability}} \\
        & MIMIC-III & VQA-RAD & Slake-VQA & Path-VQA \\
        & & Closed-ended $\mid$ Open-ended $\mid$ & Closed-ended $\mid$ Open-ended $\mid$ & Closed-ended $\mid$ Open-ended $\mid$ \\
        & Rouge-L$\big/$BLEU$\big/$F1-RadGraph & Acc $\mid$ Acc$\big/$Recall $\mid$ Acc$\big/$BLEU-1$\big/$F1 & Acc $\mid$ Acc$\big/$Recall $\mid$ Acc$\big/$BLEU-1$\big/$F1 & Acc $\mid$ Acc$\big/$Recall $\mid$ Acc$\big/$BLEU-1$\big/$F1 \\
        \hline
        BioMedGPT-10B & & & & \\
        Med-PaLM M-12B & 29.45/12.14/31.43 & $\mid$ $\mid$ /64.02/50.66 & $\mid$ $\mid$ /90.77/86.22 & $\mid$ $\mid$ /68.97/57.24 \\
        Med-PaLM M-84B & 31.47/15.36/33.96 & $\mid$ $\mid$ /69.38/59.9 & $\mid$ $\mid$ /92.7/89.28 & $\mid$ $\mid$ /70.16/59.51 \\
        Med-PaLM M-562B & 32.03/15.21/34.71 & $\mid$ $\mid$ /71.27/62.06 & $\mid$ $\mid$ /91.64/87.5 & $\mid$ $\mid$ /72.27/62.69 \\
        Visual Med-Alpaca & & & & \\
        BiomedGPT-S-33M & & 57.8 $\mid$ 13.4/ $\mid$ 40.1// & 73.3 $\mid$ 66.5/ $\mid$ 69.2// & 84.2 $\mid$ 10.7/ $\mid$ 47.6// \\
        BiomedGPT-M-93M & & 79.8 $\mid$ 53.6/ $\mid$ 69.4// & 86.8 $\mid$ 78.3/ $\mid$ 81.6// & 85.7 $\mid$ 12.5/ $\mid$ 49.2// \\
        BiomedGPT-B-182M & 30.7//31.2 & 81.3 $\mid$ 60.9/ $\mid$ 73.2// & 89.9 $\mid$ 84.3/ $\mid$ 86.1// & 88 $\mid$ 28/ $\mid$ 58.1// \\
        \hline
        Qilin-Med-VL & & & & \\
        LLaVA-Med(BioMed/CLIP) & & 83.09 $\mid$ /64.75 $\mid$ & 86.78 $\mid$ /87.11 $\mid$ & 91.09 $\mid$ /39.6 $\mid$ \\
        XrayPULSE & & & & \\
        XrayGLM & & & & \\
        \hline
        \end{tabular}
\end{threeparttable}
\end{sidewaystable}

\begin{sidewaystable}[!htp]
\footnotesize
\begin{threeparttable}
    \caption{Summary of academic medical multimodal LLMs in medical report generation and image classification ability.}
    \label{tab:m-llms3}
    \centering
        \begin{tabular}{ccccc}
        \hline
        \multirow{4}{*}{} & \textbf{Medical Report Generation Ability} & \multicolumn{3}{c}{\textbf{Medical Image Classification Ability}} \\
        & MIMIC-CXR & MIMIC-CXR & PAD-UFES-20 & CBIS-DDSM\\
        & & & & Mass $\mid$ Classification \\
        & Micro/Macro-F1-14$\big/$Micro/Macro-F1-5$\big/$F1-RadGraph$\big/$BLEU-1/4$\big/$Rouge-L$\big/$METEOR$\big/$CIDEr-D & Macro-AUC/F1 & Macro-AUC/F1 & Macro-AUC/F1 $\mid$ Macro-AUC/F1 \\
        \hline
        BioMedGPT-10B & & & & \\
        Med-PaLM M-12B & 51.41/37.31/56.54/50.57/25.2/30.9/10.43/26.16//23.43 & 76.67/38.33 & 95.57/78.42 & 70.11/47.23 $\mid$ 81.4/67.86 \\
        Med-PaLM M-84B & 53.56/39.83/57.88/51.6/26.71/32.31/11.31/27.29//26.17 & 78.35/36.83 & 97.27/84.32 & 73.09/49.98 $\mid$ 82.22/63.81 \\
        Med-PaLM M-562B & 51.6/37.81/56.28/49.86/26.06/31.73/11.5/27.49//25.27 & 79.09/41.57 & 96.08/77.03 & 73.31/51.12 $\mid$ 80.9/63.03\\
        Visual Med-Alpaca & & & & \\
        BiomedGPT-S-33M & ///////23/13/12.8 & & & - $\mid$ - \\
        BiomedGPT-M-93M & ///////23.2/13/12.9 & & & /18.7 $\mid$ /18.9 \\
        BiomedGPT-B-182M & ///////28.7/15.9/23.4 & & & /57.2 $\mid$ /72.8 \\
        \hline
        Qilin-Med-VL & & & & \\
        LLaVA-Med(BioMed/CLIP) & & & & \\
        XrayPULSE & & & & \\
        XrayGLM & & & & \\
        \hline
        \end{tabular}
\end{threeparttable}
\end{sidewaystable}

\end{document}